*Research Article*

# Smart Transformation of EFL Teaching and Learning Approaches

**Md. Russell Talukder**

Jazan University, Saudi Arabia
russell@jazanu.edu.sa

**Abstract:** The calibration of the EFL teaching and learning approaches with Artificial Intelligence can potentially facilitate a smart transformation, fostering a personalized and engaging experience in teaching and learning among the stakeholders. The paper focuses on developing an EFL Big Data Ecosystem that is based on Big Data, Analytics, Machine Learning and cluster domain of EFL teaching and learning contents. The framework has been developed on the basis of the theory that machine learning algorithms, when exposed to structured or semi-structure data stored in the cluster domains of EFL Big Data ecosystem, can cull out the patterns, similarities, and differences existing in the contents of the domains. Later these machine learning algorithms can apply these already identified patterns to perform new tasks on open Big Data platform and identify similar contents to be stored in the respective cluster domain of EFL Bigdata Ecosystem without being supervised. Accordingly, the paper uses two membranes to construe its framework, namely (i) Open Big Data Membrane that stores random data collected from various source domains and (ii) Machine Learning Membrane that stores specially prepared structured and semi-structured data. Theoretically, the structured and semi structured data are to be prepared skill-wise, attribute-wise, method-wise, and preference-wise to accommodate the personalized preferences and diverse teaching and learning needs of different individuals. Within the machine learning membrane, the paper includes a number of stages such as knowledge building, development of cluster domain of the EFL contents, integration of skill-wise cluster domain with the CEFR attribute-wise teaching and learning approaches, machine learning of the personalized preferences, resonating, machine learning of the cluster domain for proximity development and sustainable operation. The ultimate goal is to optimize the learning experience by leveraging machine learning to create tailored content that aligns with the diverse teaching and learning needs of the EFL communities. Developing a prototype following the framework exerts the potentials to provide an 'alternative to methods', transforming the process of learning into a process of acquisition.

## 1. Introduction

Methodologies and technologies, according to a recent research recommendation [1] for the institutions and teaching faculties, be directed to improve student engagement in their institutions [1]. Because, engaging students with academically purposeful activities [2] is an impactful factor for student learning and development [3-7], failing which results in growing dissatisfaction, negative experience and dropping out of school [8-10]. Being able to engage the students in the process of  learning English as a Foreign language (EFL) or as a Second Language (ESL), according to researchers, potentially transform the process of 'learning' [11] into a process of 'acquisition' [11]. The best teachers in any field are not dogmatic but usually eclectic [12], preferring to explore the best features of different methodologies to accommodate the teaching and learning needs of the students in the classroom environment. The integration of technologies in the process can "help activate and develop" [13] their sense of plausibility. Though, according to Garrett, the integration of technology, i.e., which technologies are chosen for EFL teaching, remains an issue [14], we have seen a substantial use of technologies in EFL teaching all over the world over the last three decades. The use of multimedia in problem based teaching has enhance EFL teacher's strategic capability [15] to maintain "fidelity" [15], "representational richness" [15], "time and timelessness" [15], and "individualization" [15]. In fact, the use of internet and applications run on mobile devices has blurred the boundaries of formal and informal learning, classroom -room based learning and out-of-class-room learning [16]. It has created an unprecedent opportunities for the EFL teachers and learners [17]. The use of virtual reality (VR) to facilitate embodied cognition has enabled the EFL learners to experience physical and imagined embodiment that are direct, surrogate and augmented [18]. Again, though the development of educational robots is yet at its premature stage [19], the application of robot in teaching and learning has gained much attention of the researchers [20,21]. In the recent years, the catalytic prospects of Big Data [22] and Analytics have encouraged the researchers to concentrate on 'education data mining' (EDM) and learning analytics (LA). The application of these disruptive technologies or artificial intelligence (AI) has calibrated the existing technology to detect patterns in large volumes of data. These analytical findings help the institutions to measure, collect, analyze, and report data about learners and their context. Such a data driven approach also signals a paradigm shift which signifies the necessity of AI-powered analytics for the institutions to optimize learning and understand its environment. Again, computer-assisted language learning (CALL),  EFL teaching and learning websites, software for error analysis and feedbacks for  natural language processing (NP), concordance and corpus analysis, [23] ,etc. all signify that the integration  of technologies fosters "language awareness and focuses on form" [24] in all the four modules of English.'.

Many EFL teachers are obsessed with their search for an ideal method of teaching English. As no method is elastic and versatile enough in its modalities to accommodate the diverse teaching and learning needs [13], they often end up with resentment and dissatisfaction. Over the years, there have been adequate researches advocating the "benefits" [25, 26] of following certain "methods" [27] in teaching English. On the other hand, the criticisms [28-30] emanating from the limitations of different teaching methods have also been widespread, raising doubts whether methods are altogether necessary to follow or not. In fact, volumes of journals, researches and books are showcasing the arguments both in favor of and against the use of methods.

The advent of AI with its catalytic potential has triggered a possibility of technological overhauling of the existing EFL teaching and learning approaches. Of course, it sounds altogether a new approach, leveraging a paradigm shift from the traditional methods to smart method. Such an approach, being an alternative to methods, can benefit the teachers and learners with the strengths and modalities of different methods [27] and learning approaches. It can capacitate the teachers to switch from one particular method to another, if it appears exhaustive in accommodating the teaching and learning needs.

A technological overhauling of the existing pedagogical methods is not an easy task, as it leapfrogs to many challenges. For examples, the strategies to harness the challenges and limitations of different methods using disruptive technologies such as Big Data, Analytics, and Machine Learning are absent in the field. The research gaps on how to develop a framework of AI-powered smart approach for EFL teaching and learning are quite noticeable. There is an inadequacy in the scholastic directions to leverage the use of multiple methods and AI in a regular English class. Besides, the modalities to build contents to give a personalized and engaging experience in teaching and learning using different methods simultaneously have hardly been studied.  For example, if a teacher needs to ensure that the students (i) use a particular language form correctly, (ii) understand that particular rule correctly, (iii) apply it in other communicative social settings appropriately, and (iv) develop confidence and fluency by using it in real-life settings, he or she needs to follow a number of teaching methods in a single class. First, he or she needs to adopt  and switch from one method to another, starting form audiolingual approach  [31] to grammar translation approach  [31], from communicative language teaching [31] approach to  finally humanistic approach [32].  Now the processes in which students learn are different and the modalities of different teaching methods have their specific content-requirements. So, the teacher needs to engage the students in drilling correct language and chanting [33] activities while following audiolingual method. Then he or she needs to engage them with practicing similar structures [33] when she uses grammar translation method. Next, he or she needs to engage them with dialogues and role-playing [33]activities in lifelike situations in communicative method. Finally, he or she needs engage them with personalized learning [33] activities while following humanistic approach. In response to the efforts taken by the teachers, the EFL learners subsequently "mirror and imitate the examples" [33], assume to "become a linguist in understanding the rule" [33], "stimulate and practice the language for real-life settings outside the classroom" [33]and negotiate personal learning goals with the teachers [33]. Practically, there is a big research gap as to how to prepare AI-powered EFL teaching and learning materials following the modalities of different methods and the diverse needs of the teachers and leaners.

A language class usually consists of students who reflect differences in their academic abilities. There are differences in their areas of interests as well. So, when a teacher attempts to offer a personalized experience [34]' in learning, he or she needs to accommodate the "diverse learner needs and expectations" [34], aligning the teaching contents with the interest areas and the competence-level of the students. The teacher may prefer to follow 'humanistic learning approach' for the struggling learners, 'task-based learning approach' for the proficient ones, and 'direct acquisition' or 'autonomous learning approach' for advanced ones. Moreover, in order to transform 'learning' into a process of 'acquisition', the teacher needs to opt for 'inputs [35]' that are "slightly beyond their current level of competence"[35,36].Using precision education techniques of personalized learning, including learning analytics (LA) and adaptive learning software, the EFL teacher can now accommodate all these learning needs [34]. The study is about how creating an EFL Big data ecosystem, with its catalytic prospect, can accommodate the diverse and enormous teaching and learning needs of the EFL communities to give them a personalized experience in teaching and learning ESL or EFL.

## 2. Problem Statement

In face of advancement of Information Technology (ICT) and the widespread use of internet in the recent years, the pedagogical domain of EFL or ESL has seen not only a "decline of methods" [28] but also an absence of major voice [37] from the part of the EFL teachers in the selection of technologies for teaching language. It's mainly because, the" specific teaching methods as overtly prescriptive" [29] have been inapplicable to the enormous diversity of learners and learning needs [28] in divergent learning contexts [29, 38]. Indeed, 'there is no best method' [13], though are "30 language teaching methods" [27] including Direct Method (DM), Communicative Language Teaching (CLT) Method, Task-Based Learning (TBL), Audiolingual Method (AL), Total Physical Response (TPR), and the Eclectic Approach. A critical overview of the history of pedagogical methodology of EFL or ESL provocatively nudges that" method is dead" [39, 40] and that method has only limited and limiting impact on language learning and teaching [39, 40]. The harsh criticisms against the necessity of following a method in teaching in the post-method period have led many scholars to conclude that what is needed is not an alternative but an alternative to method [39-41].

On the other hand, the domains of pedagogic methodologies have their own flagships, advocating and claiming the efficacies and usefulness of different methods. Methods, according to De Bell, are not dead and EFL teachers are aware of their usefulness and the need to go beyond them [25]. CLT's dominance during the 1980s, with its benchmarking – negotiation, interpretation and expression- almost overshadowed the existence of other methods in practice. But as soon as its limitations came into light, TBL took over. TBL has integrated the four skills of English language and connected to psycholinguistic learning, creating a new approach to teaching and learning with its problem-solving 'tasks' [26].

A new analogical perspective emerges from this debate among the methodologists and the anti-methodologists. The former favor the use of teaching methods, whereas the latter are the critics who think the scopes of the methods are exhaustive. Who is gaslighting whom is a matter of further research or endless debate. The study doesn't focus on finding on the supremacy or ineffectiveness on any methods, rather it tries to accommodate the modalities of all the methods by calibrating them with Artificial Intelligence such as Big Data, Analytics and machine learning. The incorporation of these disruptive technologies in building EFL Big Data Ecosystem, according to the hypothesis of this study, will facilitate smart transformation of the pedagogical methods and learning approaches. It will create a dynamic platform for those who find the use of methods exhaustive. At the same time, it will give a personalized experience in teaching and learning for those who find methods useful. Such smart transformation *per say* is an" alternative to methods" [13] , capable of accommodating all the diverse needs of the EFL teachers and learners.

The challenge in developing Bigdata ecosystem for EFL teaching and learning starts with the newness of big data [42]. It highlights the fact that there is not a long history of knowledge of how to teach the right skills [42]. But the newness cannot be an excuse. Many researchers argued that the new types of science need to adopt active learning techniques [43].

Evidently, the process of developing EFL Big Data ecosystem for an institution is going to be challenging unless the drive is substantiated by the right kind of technologies with the collaboration of EFL teachers and skilled data scientists or data engineers. It's mainly because of the complex nature of Bigdata with regards to its form, types, structure, source, and producer [44]. The institution needs to have its clearly defined data policies and objectives of using these data.

Another challenge emerges from the volumes of available and readily accessible EFL learning materials. There is indeed an unfathomable volumes of resource materials for EFL teaching and learning. For example, the phenomenal use of English language in this age of Industry 4.0 and globalization has created opportunities for everyone to collect suitable, interesting and engaging teaching and learning materials from a wide range of sources beyond the contents and pre-designed activities of the text books. The rise of internet and social media has made the entire society data driven, giving free and open access to millions of websites, blogs, reports, magazines, newspapers, and research publications. Social media alone produces an exponential volume of data exceeding 2.5 billion gigabytes daily [45], contributing to deepening the challenges. Most of these data are in English language covering a wide range of areas of interest of the users in their personal, social, political and economic domains. Next, the volumes of academic data produced every semester by hundreds of EFL teachers and learners with regard to teaching materials, classroom interactions, classwork, collocation drills, assignments, presentations, group discussions, quizzes, formative tests, mid-term exams and course final exams are equally enormous as well. Because of these pragmatic challenges, the systematic mining of suitable data is essentially important for building EFL bigdata ecosystem.

Further to that, if the institution is unable to customize, analyze and read data correctly, it will categorically limit its agility in addition to failing it to respond to new insight. In this connection, the technical supports of the data scientists in developing the customized teaching materials for the EFL teachers can be a panacea, but a language institute hasn't seen any such collaboration. There is a research gap to address the nature of such collaboration. There are also research gaps in other areas such as the types of new roles and responsibilities the teachers need to deliver as the domain advisors, the ICTs trainings the teachers must undergo, the logistic supports and technological integration the institute must carry out, the policy and the transformation roadmap the institution must adopt, and so on.

Giving a personalized experience in teaching and learning is only possible when these issues, complexities, modalities and limitations of methods and technological spectrums are addressed in developing an actionable framework of EFL bigdata ecosystem. There is a research gap in this particular area too as to how an institute can build up its own EFL Bigdata ecosystem and EFL data interoperability model with regards to collecting, developing and sharing materials to offer a personalized experience for the EFL teachers and learners. The study strives to demonstrate how the integration of big data analytics and ICTs in teaching domain and in developing EFL teaching materials can indeed mobilize 'innovation, creativity and smart transformation [44]' - the underlying factors of smart institution.

## 3. Objectives and Scopes of the Study

The study aims at fostering a personalized teaching and learning experience for the EFL teachers and learners both at onsite and online modes. Developing a prototype using the proposed theoretical framework will eventually facilitate a smart transformation of EFL teaching and learning approaches and minimize the limitations of different EFL teaching and learning approaches. The research will also enable the stakeholders to understand the modalities of different methods, their content-requirements, and the types of activities, tasks and exercises to be followed while using these methods. The scopes of the study cover issues including but  not limited to (i) building framework for implementation of  EFL Bigdata

Ecosystem holistically; (ii) mapping out infrastructural requirement; (iii) knowledge building; (iv) evaluating the modalities of popular EFL teaching and learning approaches; (v) modalities, activities, tasks, and objectives of skill-wise, attribute-wise, method-wise, preference-wise cluster domains; (vi) machine learning of the cluster domains; (vii) charter of collaborative partnership between EFL teachers and data scientists; (viii) EFL data interoperability model; (ix) resonating the framework of cluster domains with personalized preferences and (x) machine learning of the framework for the zone of proximity development.

## 4. Methodology

The philosophy of the research is 'deductive [46]' and 'interpretivist'. The domains of the topic cover multidisciplinary areas. The logical use of existing theories and processes [47] of machine learning, bigdata, personalized preferences for subject of learning, attribute-wise contents and the features of the modalities of the EFL teaching and learning approaches provide the basis of the theoretical framework. Depending on 'qualitative analysis' [48], relevant data were collected from literature review, surveys and interviews with the data consultants and EFL teachers. A total number of 181 research publications, books, websites, and blogs were studied to extract qualitative data. Most of these researches were carried out in the countries having higher exposure to technology-enhanced education. The study uses seven widely used and popular teaching and learning approaches of the last one hundred years to build its 'cluster domain' in the framework of EFL bigdata ecosystem. The cluster domain shall carry structured and semi-structured data or contents that are to be prepared skill-wise, attribute-wise, method-wise and preference-wise. The underlying rational is to address the EFL teaching and learning approaches from (a)" CLT" to "TBL", (b) method-based pedagogy to post-method pedagogy, and (c) from systemic discovery to critical discourse" [38]. The sources of secondary data include books, publications, reports, websites, and blogs. The sources of primary data include survey and interviews with data consultants and EFL teachers. The information search has been categorized following the objectives and scopes of the research. Accordingly, the modalities of the teaching and learning approaches, content development, framework for technological integration, machine learning, Big Data and analytics, collaborative adoptability approach for the sustainment of the paradigm shift, charter of new responsibilities, knowledge building and policy formulation guidelines have been given priority.

## 5. Theoretical Framework for Building EFL Bigdata Ecosystem

The study coins the approach as 'smart approach' as it can potentially facilitate smart transformation of EFL teaching and learning approaches, generating a personalized experience of teaching and learning. The installation of the framework practically will transform the traditional Departmental library for English language books on the data ecosystem and on the smart devices of the teachers and learners. The conceptual framework is based on big data, machine learning and personalized EFL teaching and learning contents. It is built on the idea that machine learning algorithms, when exposed to structured and semi-structured data such as specially designed EFL contents of the cluster domains, can recognize common patterns and similarities in them. Later, these algorithms can search for similar contents from the bigdata lakes to add to EFL Bigdata ecosystem.

Accordingly, the framework for EFL bigdata ecosystem has two source membranes, namely open Big Data Membrane and Machine Learning Membrane. The open Big Data Membrane consists of random data collected from various source domains and the machine learning membrane consists of the structured and semi-structured data that are to be prepared skill-wise, attribute-wise, method-wise and preference-wise. With regards to choosing level of difficulty of the EFL contents, the attributes of Common European Framework of Reference (CEFR) scale have been adopted, Hence, the EFL contents on a particular topic will have six different versions that correspond to the six different levels (A1, A2, B1, B2, C1 and C2) of difficulty in CEFR scale. Next, to facilitate personalized experience in teaching and learning, these EFL contents are further aligned with the interest areas and the modalities of the popular teaching and learning approaches. Each of these versions refers to a separate and independent cluster domain. When machine learning algorithms will be able to detect and identify patterns, trends and similarities existing in the contents of these cluster domains, they can use them automatically to identify, retrieve and collect similar contents from the open Big Data membrane and add those contents to the cluster domain they correspond to, making each of the cluster domain really big in volume, veracity and variety over the period of time. The sources of open big data membrane are mainly but not limited to books, news contents, online contents, exam papers, blogs, websites, social media, library books in hypertexts, online newspapers, online magazines, EFL researches materials, journals, seminars, classroom assessments, organization heads, EFL teachers, EFL learners, content developers, and data scientists. The machine learning membrane opens with knowledge building which refers to the study of the types of pre-determined sets of data to be included in cluster domains. It also includes the study of different theoretical, pragmatic and logistic requirements holistically. Secondly, it includes machine learning of skills, which refer to ways of teaching the machines to recognize skill-wise and attribute-wise contents. Thirdly, there is the integration phase. It refers to the ways of teaching the machine to recognize the contents that align with the modalities, tasks, and activities of the widely used EFL teaching and learning approaches. Fourthly, it's about personalized preferences. It refers to teaching the machine to categorize the areas of interest of the EFL learners. Fifthly, there comes the task of resonating. It refers to teaching the machine to map out the EFL teaching and learning needs of the students with their personalized preferences. Sixthly, it's about cluster domain for proximity development, which refers to teaching the machine to identify and build the zone of proximal development for each EFL learners. Finally, it includes sustainable operation phase. It is about streaming customized EFL teaching and learning materials from EFL bigdata ecosystem and from Big Data Open Membrane, about giving and receiving feedbacks, about troubleshooting, about collaboration of data scientists and about ensuring continuous upgradation.

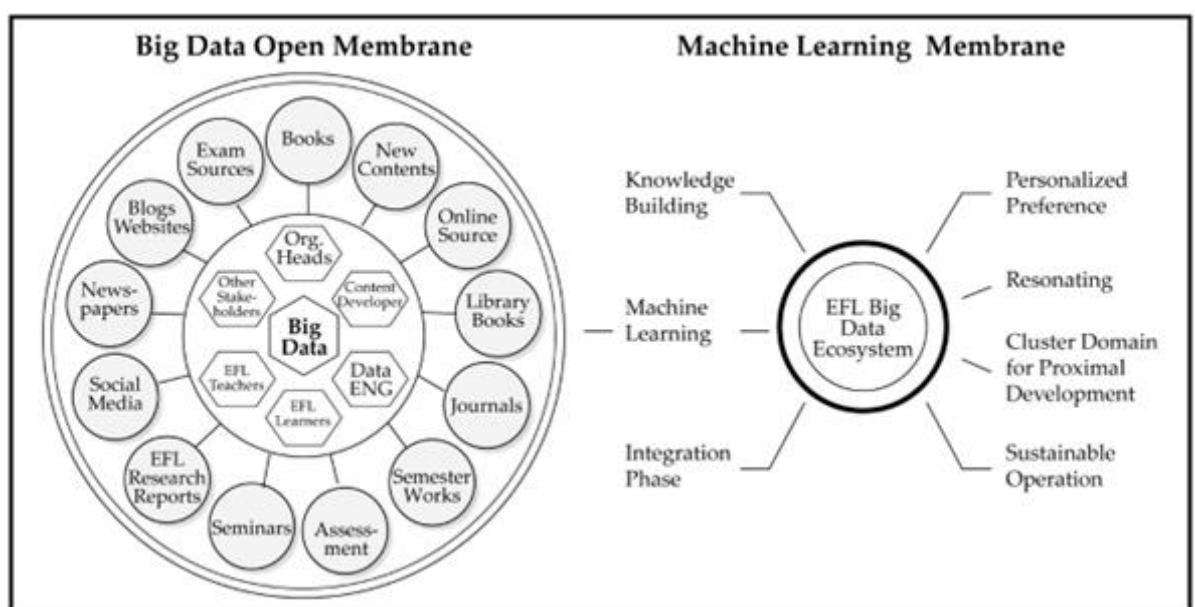

**Figure 1.** Theoretical Framework for EFL Big Data Ecosystem

## 6. Literature Review

To simplify the innate complexities of the problem emanating from its diverse areas, the literature review has been segmented subject-wise, without compromising their resonating aspects. The phases of literature review include (i) prospects and possibilities of big data and analytics, (ii) framework of integration of bigdata and analytics with the institution, (iii) study of the modalities, strengths, limitations, and contents of the selected EFL teaching methods, (iv) study of the learning approaches of the EFL learners and acquisition theories , (v) study of the personalized areas of the interest of the learners in general and (vi) machine learning of the contents  and smart transformation.

### 6.1. Big Data and Analytics

Developing institution's own Bigdata ecosystem and analytics for specific purpose of teaching and learning is possible by collecting data from the students and teachers [49].  These data need to be stored by third parties in administrative systems [49].  However, today the sources of data are not just limited to teachers and students anymore, rather they include a larger domain of producers [44] with different types [44] of structured, semi-structured and unstructured data [44]. Starting from the online materials, assessment data collected from student monitoring systems, student evaluations, classwork, semester work, online learning environment [49], books, newspapers, magazines, publications, social media, screen plays, researches, reports, publications, websites, blogs, to many other sources, the availability of data is never ending *per say*. These sources often become potentially useful when it is about developing interesting and engaging contents for teaching and learning English. Earlier teachers and students needed to visit library for collecting teaching and learning materials. But now the advancement in Information Technology, especially disruptive technologies [50], internet [51], digital world and new media [52] ,has dismantled the boundaries of time and space. They have propelled changes across the globe" [52]. With careful handling of issues such as privacy, security, ethics [53], responsibility, availability and quality of data [49], the potential of big data for education is impactful. In fact, big data has increasingly been recognized [18] for its ability to bring a qualitative change in the field of teaching and learning.

By definition Big Data refers to any digital data sets that are too large and too complex. Usually, they are generated too fast to be processed by humans or ordinary software [54]. According to Elizabeth D. Liddy, Syracuse University, big data is "a very large collection set of either textual or numeric data and, increasingly, image data …. the clear goal is to reuse it". Sometimes, the data sets may be flawed or polluted by significant amounts of noise or irrelevant information. Further, they may carry wrong information, or inflict security, or violate ethical or privacy issues, they can yet unlock meaningful and actionable information [55, 56]. In fact, Big Data offers great potential for humanitarian practices [55, 56]. Here comes the significance of analytics.

Big data analytics can exert significant impact upon an organization, when it is understood well through linking it well with the core assumptions of the organization [57]. However, data itself only provides the foundation of knowledge. According to knowledge management and hierarchy [58], data itself cannot give any insightful information for decision making and planning. In order to transform data into insightful knowledge, the organization has to ensure" the combination of experience, cognition, and competence" [59]. Then it has to focus on the use of the analytics and algorithms. Because, analytics, as they give 'knowledge of patterns in data' [58], can effectively be used for improving education, especially in the "cognitive domain, affective domain, and psychomotor domain" [60] of the EFL learners. Analytics, with the application of algorithms, can cull out trends and patterns from huge data lakes, retrieving critical information about our engagement, preferences, sentiments, and other behavioral aspects in society [61]. Analytics can transform the high volume, velocity and variety of Big Data into value [62]. In other words, it is an embracing technology that has the capacity to search, aggregate, and cross- reference large data sets [63]. Analytics can process, filter, evaluate, examine and cull out trends and patterns from structured and unstructured data [64]. In other words, the universities or organizations can use Big Data analytics for organizing data, interpreting and finally applying processed data to decisions [65].

International IT giants, multinational companies, and corporate giants worldwide are using analytics to improve their customer service and to attract new customers [61]. Besides, it is frequently used in business, and healthcare, and the consumers are increasingly aware of this [61]. On the flip side, Ellen Rose in her article, referring to Christensen and Eyring, stresses that universities that don't adapt to this new reality by reengineering themselves are committing slow institutional suicide [66].

### 6.1.1. Industrial Application of AI, Analytics and Big Data

The application of Artificial Intelligence and its associated branches including Blockchain, Internet of Things (IoT), Big Data and Machine Learning has already appeared as a game changer in different industries including aviation industry, making the air travel experiences better, personalized and rewarding for the constantly changing passengers [67]. Maintenance drones, wearable technology, electronic flight bags and document management software, for examples, have emerged as very useful in today's aviation industry. Like other service industries, the aviation industry also produces a massive amount of data everyday within its operational, management, and service domains, requiring it to take smart decision. Often these decisions are fragmented and siloed both within the organizations and outside of it. AI is being used to create an open data analytics platform to make better data-driven decisions as the industry pushes forward [68] . It has contributed to improve performance in different areas such as airport management [69], dynamic revenue management [70] and airline safety and security' [71]. The use of AI in retaining passenger loyalty [72], feedback analysis, fraud detection, catering, inflight sales, crew management, better customer experiences, fuel use optimization, predicative analysis of ticket sales and pricing has also increased substantially. AI is pacing up fast. According to Jean-Marc Cluzeau, Principal Advisor to the European Union Aviation Safety Agency (EASA), the widespread use of AI is because of its computing power, powerful algorithms and architecture and capability to process and store massive volumes of data. Machine learning with its powerful algorithms helps monitor the performance of the systems, does predictive maintenance and detects anomalies early and increases the resilience of the system" [73].

### 6.2. Integration of Bigdata and Analytics with the Educational Institution

According to SAS, a trusted analytics platform, the value of Big Data and analytics lies on how the technology is being used [64]. It can help the government, management (in education as well), and private sectors improve operational efficiencies, optimize product development, make smart decision, drive new revenue and create new opportunities for growth [64]. According to Danah Boyd and Kate Crawford, the integration of Big Data and analytics relies on 'technology', 'analysis' and 'mythology' [74]. Analytics maximizes computation power and algorithm for gathering, comparing, linking and analyzing data sets [74]. Analysis, using the large data sets, enables to identify patterns to make economic, social, technical and legal claims. And mythology is about finding a higher form of intelligence and knowledge to generate insight that were previously impossible, with the aura of truth, objectivity and accuracy [74].

Khan's theoretical framework for E-learning platform identifies eight dimensions, which is a further elaboration of Danah Boyd and Kate Crawford theoretical basis. These dimensions include (i) pedagogy; (ii) technology; (iii) interface design; (iv) evaluation; (v) management; (vi) resource support; (vii) ethics and (viii) institutional challenges. Pedagogy covers institution's content design, delivery and implementation process with strong focus on the learners' needs and objectives. Technology refers to hardware, software, learning management system (LMS), infrastructure planning, tools for communication, institution's goals and objectives, and creation of learning environment. Interface design refers to web design, navigation, accessibility and overall look and feel of the system. Evaluation refers to assessment of the learners, evaluation of the instruction and learning environment, contents, planning, institutional levels, quality of data, producers and analysis of data. Management refers to stakeholder's involvement in the learning program and its role in updating and upkeeping the learning environment, quality control, budgeting, staffing, security, scheduling, technological support providing and data collecting. Resource support covers technical and human resource support in student services and web-based technical support. Ethics is about addressing ethical issues related to plagiarism, social and political influence, diversity, bias, digital divide, privacy, information accessibility, moral and legal issues and the use of big data with the anonymity of the sources of data where required. Institutional challenges refer to administration, academic affairs, admissions, policy, accreditation, information technology services, faculty support services, class size, and data ownership [61].

Realizing the potential of Big Data and analytics in 2013, the President's Council of Advisors in Science and Technology (PCAST), America, recommended for building "a national center for high-scale machine learning for growing education data sets [75]. The council also recommended the development of competitive extramural grants to accelerate the improvement of educational materials and strategies to lead to customizable curricula for different types of students [75]. Massive Open Online Courses (MOOCs) could capture massive amounts of real-time data. As these data were pivotal in expanding research opportunities and in learning deeply about the differences, the Executive Office of the President, 2013, confidently decided to invest further in it. Notably, the administration of White House Officer of Science and Technology too continued to support the Department of Education's Institute of Educational Sciences (IES) to build the Virtual Learning Laboratory to explore the use of rapid experimentation and big data to discover better ways to help students master important concepts in core and academic subjects [76].

On the other hand, in the private sector researchers are looking to explore whether similar techniques are applicable in education [77]. The goal, like other service providing companies, is to provide a more interesting experience in learning [77].Two particular areas are given attention to namely: 'Educational Data Mining (EDM)' and 'Learning Analytics (LA)" [23]. EDM deals with application of computerized methods to detect patterns in large collections of educational data [78].  LA, as in the case with other domains, is dedicated to measure, collect, analyze and report data about learners and their context. The purpose is to optimize learning and environments where it takes place [23]. LA and EDM are constituent components of Big Data Ecosystem [23].

Such changes in learning domains essentially require the educational institutions to adopt a new kind of strategic framework. In other words, it necessitates the inclusion of data engineers or data scientists and EFL teachers in the same loop of the organizational and human capacity [22].  The collaboration of data scientists and EFL teachers in the use of big data [22] can contribute to organization's human capital development. In fact, the changes highlight the need of training and professional upskilling [22], investment in organizational infrastructure [22],  partnership building among the stakeholders [22], and formation of multidisciplinary teams [22]. The collaboration facilitates      scientific research and innovation [22] and strategic sustainment in onsite and online modes of teaching. The effectiveness of the disruptive technology depends on how the EFL teachers and data scientists have used the Big Data tool for data visualization, predictive analytics, and EFL materials designing. It also largely depends on how the knowledge of patterns [79] in the learning behavior of the EFL learners, statistics, communication, and domain expertise of the EFL teachers are used in their attempts to content development for EFL bigdata ecosystem.

### 6.3. Study of the Modalities, Strengths, Limitations, and Contents of EFL Teaching Methods

Different methods have significant impacts in achieving teaching goals for EFL teachers. People have different preferences for communication styles, which itself reflects their values and which may create programming problems. The preference for a style of programs which is low in ambiguity and in difficulty may carry over into all other communications [80]. India, often called the museum of methods [81] with respect to  teaching English language, uses Reading Method, Grammar Translation Method, Direct Method,  Structural Approach, The Silent Way, Structural Oral Situational Approach, Natural Method, Suggestopedia, Audio Lingual Method, Community Language Learning, Bi-Lingual Method, Skilled-Based Learning and Teaching, and Communicative Language Teaching [82]. In Bangladesh the curriculum reform, initially at school and later on, at all levels, has been carried out to align with Communicative Language Teaching method, replacing the traditional Grammar Translation Method that prevailed till mid-1990s [83]. In China the introduction of CLT into English classroom aimed at meeting the larger demands of producing graduates with communicative competence, though it has been facing many other challenges like large class size, lack of trained teachers and budget constraints [84]. In Indonesia, according to recent research, the two methods that are predominantly in use are Grammar Translation Method and Direct Method [85]

Larsen- Freeman recommends seven methods for teaching EFL namely:  Grammar Translation Method (GTM), The Direct Method (DM), The Audio-Lingual Method (AL), The Silent Way, Suggestopedia, and Community Language [86]. Dr. Shiladitya Sen, Instructional Specialist at Montclair State University, NY, garnering Kachru's concentric model, prefers Direct Method for teaching the native speakers. Kachru's concentric model of 'three circle' focuses on the four functions of language including (i) instrumental, (ii) regulative, (iii) interpersonal and (iv) imaginative or innovative [87] functions.  Dr. Sen asserts that Americanization movement for creating a coherent, monolingual national identity [88], embedding in the pedagogy the images of an ideal nation with its homogeneous citizens to minimize the difference of immigrants' language practices [88], will fail to harness the power of diversity through pluralism [89].Such assertion strongly justifies the use multiple approaches in teaching English. Professor Tania Hossain of Waseda University, Tokyo, Japan, adds one more method to Larsen-Freeman cart, namely Content and language integrated learning (CLIL), in her attempts to identify the EFL methods used in Japan.  Of the 30 different methods [27] used in EFL teaching, the modalities, strengths and limitations of some of the widely used methods have been evaluated for content analysis. These methods cover a long period of EFL teaching history since 1900. The study is essential for determining data mining strategies, designing effective programming, building collaborative partnership and preparing effective and engaging learning contents.

### 6.3.1. Communicative Language Teaching (CLT)

CLT is an interaction-based method. Its prime goal is to enable the learners to communicate in the target language using interactions [90]. It encourages them to share their personal experiences into their language learning environment [91]. The following table shows its modalities, strengths, limitations and types of classroom activities that are followed while this method is used by an EFL teacher.

Table 1. Modalities of Communicative Language Teaching

| Methods | Modalities and Strengths | Limitations | Classroom Activities |
|---|---|---|---|
| Communicative language teaching (CLT) | (a) Focuses on negotiation, interpretation, and expression [90,92,93]; (b) Following the principles of *Speech and Act , CLT stresses* on how language users perform speech acts such as requesting, informing, apologizing, etc. [94]; (c) Incorporating interactions and socio cultural norms [95] of real-life contexts; (d) creative, unpredictable and purposeful use of language in the classroom practices for information sharing and negotiating meaning [30]; (e) Focusing on Communicative Orientation based on 'a partial simulation of meaningful exchanges that take place outside the class ' [30]; (f) facilitating innovative strategies like role 'plays, games, scenarios' [30], problem-solving [96], situational conversations, assignments, etc. and (g) Stressing on fluency and eloquence | (a) CLT's boastful claims on authenticity, acceptability, and adaptability are not validated by the data-based classroom-oriented investigations carried out by the researchers [38, 97, 91, 98, 99] (b) choked with jargons [100]; (c) serious intellectual confusion [100]; (d) over-generalizing the insights to such an extent till they become virtually meaningless [100]; (e) traditional textbooks and coursework focusing on passing the tests didn't complement communicative approach [101, 102]; (f) encourage noise, disorder, etc. that are conflicting to the norms of classroom [103] | (a) Oral 'role playing [104]' in pairs to develop communicative abilities in certain situations ;(b) Taking interviews [105] involving the learners in pairs to facilitate 'interpersonal skills [105].(c) Collaborative group work [105] engaging each student with specific tasks to facilitate interactions and creative discussion on the information gap [106] filling activities to obtain the previously unknown information;(e) Opinion sharing [106] activities for engaging the students in discussion on topics they are familiar with and (f) Mingling activities or scavenger hunt [105] activities to engage students in interactions |

### 6.3.2. Audiolingual Method (AL)

Based on behaviorist theory, AL focuses on training the learners certain traits of living beings through a system of reinforcement [107]. Table 2 shows its modalities, strengths, limitations and the types of classroom activities that align with the method.

Table 2. Modalities of Audiolingual Method

| Methods | Modalities and Strengths | Limitations | Classroom activities |
|---|---|---|---|
| Audiolingual Method | Focusing (a) on linear and additive aspect of language learning [38] (b) Modelling correct expressions [33] ([presentation) – facilitating imitation of the expression (practice) – continuing the practice till 'correct' language becomes automatic and habitual (production). Founded on 'behaviorist theory [107]. (c) Focusing on listening and speaking, the order of learning follows -listening, speaking, reading and writing – sequences (d) Use of language laboratory, visual aids, gesture, (e) similar to direct method, stressing less focus on mother tongue | (a) Expressions 'taught in chunk' [33]; (b) focused with accuracy; (d) The entire class drilling or practicing a single expression for 'ten times' [33] is self-evident of its being a slow pace method ; (e) Less space for cognitive learning ; (f) Learners having no control over their learning [108]; (g) Reading is neglected | (a) Engaging the learners in drills and pattern practice; repetition, inflection, replacement and restatement [31], (b) Engaging the students in dialogues [109] and in listening comprehension [110];(c) collocations; (d) Writing composition following the oral exercises; (e) Reading materials to be taught orally first and (f) Mimicking dialogues [110] |

### 6.3.3. Direct Method (DM)

Focusing on oral skill, DM uses only the target language. It is also known as the natural approach. It relies on question-answer pattern-based teaching [111]. The use of learner's native language is highly discouraged. Table 3 shows the types of classroom activities, modalities, strengths and limitations of the method.

Table 3. Modalities of Direct Method

| Methods | Modalities and Strengths | Limitations | Classroom Activities |
|---|---|---|---|
| Direct Method | (a) Based on the doctrine of Sauveur and Franke in 1900 that language teaching should be undertaken within the target- | (a) Difficult for slow | (a) Question and answer patterned contents; (b) |

| Or Natural Method | language system [112]. Teaching is carried in English to 'establish a relation between experience and expression [113]' to enable the students to communicate in English, and translations in mother tongue are never used [114], and 'preferences given on speaking [115]', not on grammar; (b) Students learn directly, instinctive and naturally; many ESL companies prefer teacher using this method as they just need him or her to speak in English in the class [96] and it is extensively used for online teaching; (c) Using inductive or indirect approach for grammar through the 'implication of the situation creation [116], (d)Improving fluency, making writing easier, facilitating understanding of the language and alertness and participation of the learners [113, 114], (e) Making lessons more real-world by introducing organic variables common in the cultural and locale of language use [111] | learners to cope up [114] and (b) Requiring highly proficient EFL teachers [114] | Using visuals and real-life images for teaching ideas, concepts and vocabularies; (c) Reading materials for dictation; (d) Passages, plays, stories, dialogues, etc. for engaging students in reading aloud; (e) Self-correction by the students;(f) Paragraph writing using their own words [113];(g) Engaging students in conversations, asking and answering questions |

### 6.3.4. Task Based Learning (TBL)

A few others methods including Problem-based Learning (PBL), Content-based Instruction (CBI), Content and language integrated learning (CLIL), and Task-supported language teaching (TSLT), though with significance differences, show certain level of similarities to the discourse and modalities of Task-Based Learning. For example, in CLIL, a learner learns a subject using a foreign langue. So, he learns both the language and the subject. Again, TBLT follows pre-task, post-task order, whereas TSLT follows present, practice and produce sequence [117]. Tasks given to the students have to be relevant and interesting to them. The use of authentic language is more important than answering grammar or lexical questions. Clustering the students in small groups, encouraging them to collaborate with others, giving and receiving feedbacks from the peers and motivating them to communicate with others are important check list. Table 4 shows the modalities, strengths, limitations and types of contents used in the TBL.

**Table 4.** Modalities of Task-Based Learning

| Methods | Modalities and Strengths | Limitations | Classroom Activities |
|---|---|---|---|
| Task-based learning (TBL) | (a) Engaging the students in the tasks, problem-solving, 'information gap activities, reasoning gap activities, opening gap activities and interactions [118], discussions, and presentations ; (b) Using authentic English language to complete meaningful tasks focusing on pragmatics ( meaning), information gaps and leaner's autonomy to choose linguistic resources [26], (c) Methodologically being very student-centered [119], (d) Tapping into learners' natural mechanisms for second-language [120], hence, tasks per say must have clearly defined non-linguistic outcome [121], (e) Engaging the learners in tangible linguistic products with texts, audio or video recording, presentations, review of each other works, giving feedbacks in physical classroom setting [121] | (a) use of mother tongue [122, 123]; (b) classroom management [103, 122, 124] ;(c) quality of the target language produced; (d) 'hinder language production [125]; (e) tasks designed neglecting any intent of communication [122] | (a) Pre-tasks: Presenting contents with presenting picture, audio, or video and demonstrating the task [126]; (b) modeling tasks in authentic language with information gaps, reasoning gaps, opinion gaps [118]; (c) Texts, audio or video recording, presentations, etc. to demonstrates tasks and tangible linguistic products [26] and(d) Modeling contents with focus on meaningful communication [26] |

### 6.3.5. Total Physical Response (TPR)

Using comprehension approach, the method tries to teach the target language. Learners are not required to show outputs at its early stage. The method focuses on emphasizing the importance of listening to language development in the beginning [86]. Table 5 shows the modalities, strengths, limitations and types of classroom activities used in TPR.

**Table 5.** Modalities of Total Physical Response

| Methods | Modalities and Strengths | Limitations | Classroom Activities |
|---|---|---|---|
| Total Physical Response (TPR) [127] | (a) Founded on the hypothesis that brain is neurologically predisposed to learn language through listening; and that after internalizing the target language through inputs, the learners develop speech naturally and spontaneously [127]; (b) Actions before words; teacher focusing on movements, physical expressions such as miming, gesturing, or acting out the language before or while teaching the words or expressions or ideas ; (b) Pairing physical and intellectual analysis – mining the meaning of the words enables the students to learn it using different parts of | (a) criticized for being suitable for beginner learners, especially for kids [130] | (a) Extensive and varied modular use of 'Realia' [31] such materials and objects from real life, specifications, extracts from company, industry or organization, manuals, diagrams, etc."; (b) Modelling lessons in the form illustration, pictures, posters and props linking physical |

| | their brain; (c) Effective for children and young learners for physical and online teaching ; (e) 'it's aptitude free and effective for a class of students with mixed academic abilities, especially for those who have physical disabilities [128], dyslexia and other learning difficulties [129]    (f) Enabling the leaners remember vocabulary (f) Facilitating opportunities for the learners to ventilate their  energy and stay focused for long [64] | | expression with the language [31]; (c) Modeling interactive lessons for the learners on different skills to practice privately. Asher used CD-ROM to facilitate private learning [31, 131]. |

### 6.4. Study of the Learning Approaches of the EFL learners

The nature of the nations and societies in each Kachru's concentric circle- inner, outer, and expanding [87]- reshapes global English. While developing EFL teaching and learning contents, in addition to accommodating the functions of using English in a country as illustrated in Kachru's concentric circle, EFL teachers should also algin the contents with EFL teaching and learning approaches. Because, the purpose of learning English is different, when it is compared among native speakers of the language (ENL), and those who are learning it as a second language (ESL) or   as a foreign language (EFL) [132]. This specific distinction encompasses the countries in Asia into two circles, namely the outer circle and the inner circle [133]. Mapping the teaching methodologies with the ways the learners learn a second language provides insightful information for the teachers, content developer or educationists. Because, how the learners learn a second language is as equally important as the methods that are applied by the teachers. Without understanding the different processes of learning, trying to develop a framework for developing Big Data Ecosystem for EFL learning is unlikely to overcome bias. The following table briefly summarizes the basic constructs of the different process of learning EFL.

### 6.4.1. Acquisition

Acquisition is the process in which learners gain the ability to perceive, comprehend, produce, and use words, phrases or sentences to communicate. Language acquisition refers to first-language acquisition. It is similar to the ways the infants perceive, comprehend and communicate in their native language. It can be spoken language or signed language [134]. Table 6 shows the summary of the modalities and guidelines for the teachers to develop contents for acquisition.

**Table 6.** Modalities of Acquisition approach

| Approach | Modalities | Manuals for the Teachers or content developers |
|---|---|---|
| Acquisition [135] | (a) Using 'parallel patterns and development' [33] and 'similar stages' [33] of learning that exists in one's first language; starting with simple structures in the beginning and the moving towards complex ones.  (b) Words with meaning are learnt before the grammatical words that are void of meaning like it, was, on, to, etc. [136]. | (a) Contents need to strictly follow the stages, starting from simpler stages to complex stages, in order to align with the natural order of acquisition [33]. (b) Teacher should give space for errors and facilitate learning through examples, repetition and exposure [33]. |

### 6.4.2. Contrastive Analysis

Contrastive analysis is heavily used the area of second language acquisition. It refers to the systematic study of two languages. In the context of this research, the two languages it refers to include are (i) the native language of the learners and (ii) English language. The objective of contrastive analysis is mainly but not limited to identifying the structural differences and similarities of the two languages [137]. Table 7 shows the modalities and manuals for the teachers in brief.

**Table 7.** Modalities of Contrastive Analysis

| Approach | Modalities | Manuals for the Teachers or content developers |
|---|---|---|
| Contrastive    Analysis [138] | (a) Focusing on predicting the contents that the learners will find difficult based on the comparisons of those contents with their first language [138]; (b) The discourse centering around exploring how English is similar and different from the first language [139]. | (a) Content developers or teachers need model lessons identifying the similarities and dissimilarities of English with the first language of the leaners [33]. |

### 6.4.3. Natural Approach or Humanistic Approach

Following what Krashen and Terrell postulated back in 1980s, the theory of language acquisition dominates natural approach [36]. Table 8 shows the key modalities and guidelines for content development in brief.

**Table 8.** Modalities of Natural Approach

| Approach | Modalities | Manuals for the Teachers or content developers |
|---|---|---|
| Natural Approach [36] Humanistic Approaches [140] | (a) Using the language for expressing feelings, emotions and everyday needs; (b) Staging learning with what is familiar and then moving to new or unfamiliar topics [140]; (c) Correcting errors and giving comprehensible inputs | (a) Modelling contents to expose the learners directly to English language; (b) Avoiding explicit teaching of grammar and letting it to be discovered by the learners; (c) Allowing the learners space for errors and then correct them, as with the case with humanistic approach that support and nurture child when learning first language [33]. |

### 6.4.4. Learning as a process of forming habits

The Economist, in a recent publication, argues that ongoing skill acquisition is critical to persistent professional relevance. However, persistent learning not a matter of choice, rather it is a matter of habit that results from careful cultivation [141]. Table 9 shows the modalities and manuals for content development in brief.

**Table 9.** Modalities of Learning as a Habit

| Approach | Modalities | Manuals for the Teachers or content developers |
|---|---|---|
| Learning as a process of forming habits [80] | (a) Following the ways, a first language is learnt [80]; (b) building habits of learning; (c) exposing the learners to correct models of expressions, allowing them to imitate and practice till the correct expressions become habitual [80] | (a) Modeling contents, lessons and exercises in way so that possibility of errors is curtailed to minimum level; (b) correcting the errors, the more the errors are the more the absence of habits; (c) repetition, drilling and keeping the pace of learning same for the entire class. [33], as with the case in audiolingual method of teaching |

### 6.4.5. Humanistic Learning Approach

The humanistic learning approach focuses on activities that involve leaner's feelings, intellect, social skills, artistic skills, practical skills, etc. as part of their education. The key learning elements include self-esteem, goals, and full autonomy [142]**.** Table 10 shows the modalities and guidelines for content development in brief.

**Table 10**. Modalities of Humanistic Learning approach

| Approach | Modalities | Manuals for the Teachers or content developers |
|---|---|---|
| Humanistic Learning Theories [143] [144] | (a) Learning a language, as in the case with the first language, in loving and supported setting, using it to express the core and everyday needs, feeling, and emotions [143]; (b) focusing on learners' emotional responses in the classroom as importantly as their intellectual processes [144]. | (a) Modeling the contents, exercises, lessons that require the learners to meditate, visualize or brain-storm for five-minutes before starting them in the beginning or at the end of the class, (b) teacher may voice talk during these moment of contemplation and explore an internal journey : (c) engaging the learners to record everyday event in personal journals and discuss in the class; (d) partnership building with the peers, and setting up co-monitoring cell to counsel each other in pairs [33]. |

### 6.4.6. Cognitive Learning Approach

Cognitive learning focuses on the internal and external factors on mental process that contributes to information processing and effective learning. Cognitive means the process of acquiring knowledge or the process of understanding with the help of senses, thought and experience [145]. Table 11 shows the modalities and manuals for developing contents in brief.

**Table 11.** Modalities of Cognitive Learning Approach

| Approach | Modalities | Manuals for the Teachers or content developers |
|---|---|---|
| Cognitive Learning | (a) Learner's thought, beliefs , attitude and values [146]; learning through development of thinking; learners receives | (a) Modeling contents for identifying, analyzing and explaining patterns of expressions; (b) |

| Theory | input or stimulus, processes them in mind, and then reflects or acts upon the stimulus [147]; hence the lessons given should be predominantly intellectual or they should have a role to engage the cognition process of the learners; (b) Facilitating the learners to use their reasoning and deductive abilities to recognize rules and grammar [148]; (c) Facilitating the learners to apply strategies like labelling, classifying, organizing, terminology about language [149], and (d) Mental process or the cognition process includes observation, attention, perception, organization, storing and retrieval, categorizing and generalizing | Focusing on the analysis of the written contents; (c) Comparing and contrasting the first language and second language as with the grammar-translation method; (d) Transforming learning as an intellectual activities [33] through analysis, reasoning, functional practice in naturalistic setting, and formal exercise and practice with structures and sounds [150], etc. |

### 6.4.7. Social Constructivist Approach

Social constructivist approach is embedded on the theory that human development is socially situated and that knowledge is constructed through interaction with others [151]. Table 12 shows the modalities and guidelines for developing the contents in brief.

**Table 12.** Modalities of Social Constructivist Approach

| Approach | Modalities | Manuals for the Teachers or content developers |
|---|---|---|
| Social Constructivist Approach | (a) Focusing on ideas that leaning a second language happens trough interactions with, focus guidance from and informed 'other' [140]; (b) Using what the learners already know as the starting point, the teachers use a framework for future learning goals for each and guide them to achieve feasibly step by step their potential knowledge –which is commonly known as scaffolding; ( c) Determining the zone of proximal development (ZPD) by identifying the difference between actual development and potential development; where actual development refers to learner's capability of problem-solving alone and potential development to the capability of the same under adult guidance or in collaborative setting [33]. | (a) Modelling contents, lessons and exercises on giving suggestions, praising, reminding, encouraging rehearsals, being explicit about organization, and providing partial and whole activities; [33] (b) Modelling contents that reflect parent-child situations, keeping the organization learning goals as much as possible on the same domain. |

### 6.4.8. Learner Autonomy Approach

Learner autonomy approach refers to managing one's learning by oneself. In other words, it is an independent learning approach. Learning can be facilitated without constant interactions of the classroom. One can develop skills independently from outside of the classroom [152]. Table 13 shows the modalities and manuals for content development in brief.

**Table 13**. Modalities of Autonomous Learning Approach

| Learning process | Modalities | Manuals for the Teachers or content developers |
|---|---|---|
| Learner Autonomy Approach | (a) Focusing on placing the leaners in control of their learning, as the successful leaners like to take their own learning responsibility without relying on the teachers or classroom opportunities; (b) Facilitating the learners to manage and make decision about their own learning | (a) Letting the learners decide at the start of the class what contents, lessons, exercises they would love to complete, the objectives they want to achieve and letting them plan how they are going to prepare them, formulate questions on them, answer them and achieve the learning goals; (b) teacher acts as bystander facilitating condition for learning; (c) Facilitate installing class library to help learners choose reading materials, (d) allowing learners self-access resources; (e) engaging the learner in their own projects and activities [33] |

### 6.5. Study of the Personalized Areas of the Interest of the Learners

How the learners choose their learning contents depends on a few factors. The way they give preferences for certain contents varies from individual to individual. The underlying factors include their styles of metacognition (self-regulation, self-estimation, etc.), learning habits and behaviors, ages, level of motivation, learning goals, available learning time, and amount of prior exposure to learning English language. Social media platform is a good source of learning about how a person mines his or her areas of preferences [153]. An alternative way to learn about the common areas of interest is to follow the topics, subjects and activities that are covered in the elementary, secondary and higher secondary levels of education. Again, the different subjects offered at universities for further studies at graduates and post graduate levels are also the cases here in point. These common areas of interest include Sciences, Architectures, Constructions, Business and Management, Education and Upskilling, Creative Arts and Design, Healthcare, Medicines, Computer Science and ICT, Artificial Intelligence, social media, Law and Jurisprudence, Engineering, Humanities, Travelling and Hospitality Industry, Fitness and Nutrition, and Social Studies and Media. However, it is important to convey to the students that no credit is given for selecting certain subjects or areas of interest [154]. They should be explained that the purpose behind asking them to choose their areas of interest is to offer them a personalized experience in learning. The purpose is also about offering an anxiety free environment of learning. The list of personalized preferences based on study of the contents of the books followed at different levels of education at primary, secondary and tertiary levels of education includes 'sports', 'pastimes', 'adventure', 'social media', 'history', 'literature', 'belief', 'lifestyles', 'movies', celebrities', 'culture', 'society', 'money', 'music', 'economics', 'leadership', 'science', 'technology', 'education', 'careers', 'politics', 'social issues', 'traveling', 'healthcare', 'trade and commerce', 'banking', 'fashion', 'art and craft', 'civilization', 'personalities', 'challenges and crisis', 'international issues', 'research and innovation', 'agriculture', 'city life and country life', 'marine life', 'wars and conflicts', 'planets and astronomy', etc. . The title of these topics can be used for gathering unlimited interesting learning materials from online sources.

### 6.6. Machine Learning and AI for Building Contents

Machine learning [155] refers to the study of understanding and building methods that learn or methods that leverage data to improve performance on some set of tasks. It is a branch of artificial intelligence (AI), having the systems that generate their own knowledge by culling out patterns from the raw data [156]. It can help generate solutions, when they are not known to the programmers. Machine learning can identify characteristics involving similarities and differences. Using machine learning, for example, satellite images can be grouped on the basis of the presence of land, sea or clouds.

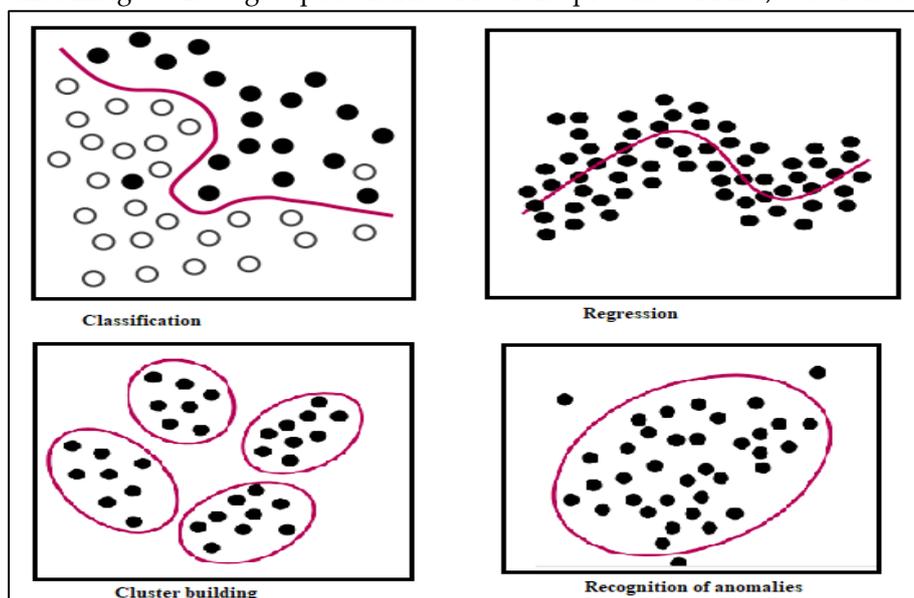

**Figure 2.** Machine learning of patterns, regressions, cluster building/Adaptation of the illustration of Nico Luck

Because of the complex nature of the processing of speech and its following coordination of actions, people don't understand it adequately. But when the machine is given to learn, it can cull out patterns and transfer them to other tasks. This version of AI is known as MPpAI, that is, "Machine Learning-powered Artificial Intelligence'. The task to be performed can be changed depending on time, the user and other parameters [157]. MLpAI can be of two forms: (i) supervised and (ii) unsupervised learning. Supervised learning refers to the form of AI that learns with the help of additional inputs from a teacher. 'Supervised learning utilizes labeled datasets to categorize or make predictions; this requires some kind of human intervention to label input data correctly [158]'. The relevant characteristics of the data are pre-defined, allowing classification, mapping subjects into image, regression or prediction and detection of anomalies [159]. and (ii) unsupervised, referring to the form of AI that can learn purely on the basis of pre-determined data. When machine learning builds clusters from unknown data, it is also an example of unsupervised learning. Machine learning works on with the help of numerous algorithms like naive Bayesian algorithms, support vector machine algorithms or different variants of decision trees. But latest development on machine learning method is deep learning, and Chat GPT is an example of how phenomenally potential it can be in generating human like texts [160].

## 7. Findings and Discussions and Building EFL Bigdata Ecosystem

The literature review denotes a number of insightful findings that can facelift the efforts of building and application of EFL Bigdata Ecosystem to make teaching and learning engaging and personalized for the EFL teachers and learners.

**Firstly**, the incorporation of Big Data into educational institutions can potentially enhance their operational efficiencies, in addition to enabling them to make smart decision, optimize production development and to create new opportunities for growth [64]. The micro domain of Big Data ecosystem in educational institution is comprised of Learning Analytics (LA) and Educational Data Mining (EDM) Membranes. EDM detects patterns in large collections of educational data [78] and LA measures, collects, analyzes and reports data about learners and their context [23]. The application the technology optimizes learning and the environments where it takes place [23]. The holistic domain of the Big Data Ecosystem in educational institution is comprised of a number of membranes. These include (i) pedagogy; (ii) technology; (iii) interface design; (iv) evaluation; (v) management; (vi) resource support; (vii) ethics and (viii) institutional challenges and to-do lists [61].

**Secondly**, analytics can retrieve critical information about our engagement, preferences, sentiments, and other behavioral aspects in society [61], which can be used to give personalized experiences in EFL teaching and learning. Analytics can help educational institutions organize, interpret and apply processed data to decisions [65]. It can transform Big Data into value [62]. It can search, aggregate, and cross- reference large data set [63]. Analytics can further process, filter, evaluate, examine and cull out trends and patterns from structured and unstructured data [64].

**Thirdly**, supervised and unsupervised machine learning and deep learning [157-158] within the framework of EFL Big Data Ecosystem can facilitate a smart transformation of the EFL teaching and learning approaches, enabling the stakeholders not only to overcome the limitations of different methods but also to facilitate them in accommodating the diverse needs of the teachers and learners. Machine learning algorithms can generate their own knowledge by culling out patterns from the raw data [156]. This particular ability can be used to develop EFL Big Data Ecosystem.

**Fourthly**, understanding about the modalities, strengths and limitations of the widely used EFL teaching methods [90, 91, 107], and learning about the types of contents used in these methods are impactfully significant. They enable the stakeholders to build engaging contents that are effective in achieving learning goals. They add a personalized experience in teaching, as the teachers can choose any method or methods that they find suitable for themselves and for the learners. The limitations of a particular method can be avoided by switching to another method that is free from those limitations.

**Fifthly**, the styles of metacognition (self-regulation, self-estimation, etc.), learning habits and behaviours, ages, level of motivation, learning goals, available learning time, amount of prior exposure to learning English language and areas of interest in real life motive the learners to mine their areas of preferences [153]. In addition to all these, the institutions require to have their data-mining and data-sharing policies, clearly defined application goals, theoretical framework and effective evaluation process. Digitizing the institutions with the disruptive technologies also requires them to examine their infrastructure, installations and human resources through new lens. The unity of purpose, determination, policies, logistics, critical infrastructures, rights and liberties, bias and inclusion, privacy and ethics and commitment to adopt Big Data and Analytics are crucial for ensuring tangible return for all the stakeholders.

**Finally,** having an agile framework supported by data engineers will ease the tasks of implementation for the EFL teachers and the institutions. The management needs to build knowledge, covering a wide range issue that need microscopic observations and analysis. A holistic EFL Data ecosystem can be panacea as it has the catalytic potential to aggregate, store and stream the learning contents for personalized and engaging teaching and learning experience.

## 7.1. Knowledge Building

In the knowledge building phase, according the findings of the study, the organization needs to identify and include the theoretical, logistics and pragmatic requirements holistically. Gathering this knowledge is pivotal in the process of smart transformation. Knowledge building covers three fundamental areas including Theoretical, Pragmatics and Logistic.

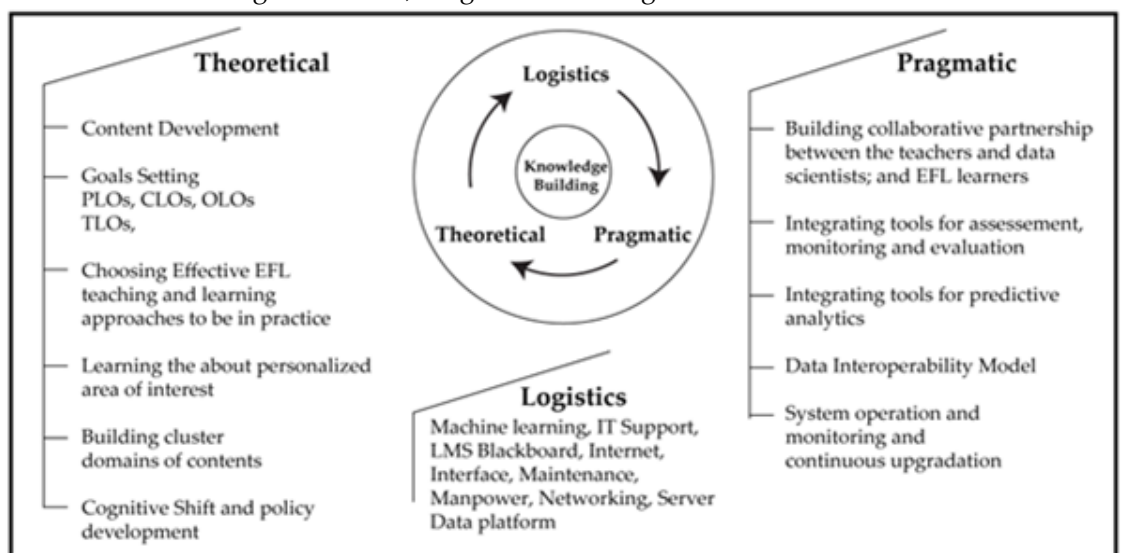

Figure 3. Theoretical, logistics and pragmatic knowledge building

### 7.1.1. Theoretical

The issues to cover include machine learning techniques[155, 156] content development [87, 132], goal settings [133], choosing EFL teaching and learning approaches to be in practices [27, 86 ,78], learning about the personalized areas of interests [153], cognitive shift, policy Development and human capital development [61].

#### 7.1.1.1. Contents Development for Cluster Domain for Machine Learning

Contents have to be developed skill-wise following the different attributes of CEFR. Based on the level of difficulties, each of the content sets of reading, writing, listening and speaking will have eight different sets including A1, A2 (Beginner), B1, B2 (Intermediate) and C1, C2 (Advance). Activities have to be designed based on the modalities of different teaching and learning methods, so that the stakeholders can choose the ones that are suitable for them. Accordingly, DM, CLT, TBL, AL, TPR, CL, and ATL will have different types of activities aligned with their modalities, strengths and tasks' requirements. The paper coins the term, cluster domain, to refer to the contents of different skills. Contents and tasks need to address the core issue of fostering the cognitive operation of the EFL learners. Research shows that the cognitive demand of writing contributes to the enhancement reading skills [161-164] as well. The contents of four skills have to be preference-wise. The skill-wise manuals for developing classroom activities for reading, writing, speaking and listening are shown in the tables below:

#### 7.1.1.2. Activities, Tools and Cognitive Challenges of Reading Contents

The topics must reflect diversities in contents that conform to the areas of interest of the leaners in real life. The list of personalized preferences of the leaners has been discussed in 6.5. under the title 'Study of the personalized areas of the interest of the learners in general [153]. Reading narratives comprised of short and long expressions and texts need to be prepared attribute-wise in CEFR scales. The modalities and strengths of different methods and approaches discussed in 6.3.1 − 6.3.5 and 6.4.1 − 6.4.8 need to work as the guiding principles in developing the contents of Reading Skill. Table 14 shows the types of activities, tools, mechanics, skills and cognitive challenges the contents of the cluster domain must have.

**Table 14.** Activities, tools, mechanics, skills and cognitive challenges for reading content

| Modules | Activities | Tools, Mechanics & Skills | Cognitive Challenges |
|---|---|---|---|
| Reading | (1) sentence completion; <br> (2) summary, note, table, flow-chart completion; <br> (3) short answer questions; <br> (4) diagram label completion; <br> (5) multiple choice, <br> (6) matching question, <br> (7) matching heading, <br> (8) matching sentence ending and <br> (9) true, false, not given questions | (1) Scanning; <br> (2) Scheming: <br> (3) Reading the for main idea of the passage; <br> (4) Contextualized reading; <br> (5) Intensive reading; <br> (6) Inferencing | (1) Identifying author's opinions <br> (2) Identifying writer's purpose <br> (3) Finding specific information <br> (4) Completing Diagram or table <br> (5) Completing summary <br> (6) Identifying the main ideas or key arguments <br> (7) Finding difference between main idea and supporting details |

### 7.1.1.3. Activities, Tools and Cognitive Challenges of Writing Contents and Tasks

Following Doyle's and Carter theoretical framework for writing the tasks can be of four categories [166]; namely (i) memory tasks (ii) procedural tasks or routine tasks, (iii) understanding or comprehension tasks and opinion tasks. Memory tasks deem to be prepared to recall or recognize information or a list of terms that the leaners have already seen before (paragraph, passage, essay, composition, summary, precis, short notes, short messages, etc.). Procedural tasks or routing tasks deem to be prepared to apply predicable formula or process to 'solve a set problem or search-and-match strategy to locate words or passages in a text [167], (describing scientific process, sequences of events, how things work, formulation, definition, classification, evolution, etc.). Understanding or comprehension tasks deem to be given to draw inferences from the information sets the students have already encountered before, and to guide the students to apply a procedure from many to solve a problem (description of statistical graphs, statistical tables, comparison, contrasting, analysis, evaluation, etc.). Opinion tasks deem to be prepared to teach and learn to forward a preference for something. (Critical appreciation, argument, analysis, analogy, reviewing). Table 15 shows the types of activities, tools, mechanics, skills and cognitive challenges the cluster domain must have for technical writing tasks.

**Table 15.** Activities, tools, mechanics, skills and cognitive challenges for technical writing

| Modules | Activities | Tools, Mechanics & Skills | Cognitive Challenges |
|---|---|---|---|
| Writing Technical | Analyzing technical or statistical data given in graphs: <br> (1) bar graph, <br> (2) linear data, <br> (3) pie chart, <br> (4) regression model, <br> (5) table <br> (6) map and <br> (7) process | (1) Analyzing Data, <br> (2) Describing the features <br> (3) defining important terms <br> (4) exemplifying <br> (5) classifying <br> (6) comparison <br> (7) contrasting <br> (8) inferencing <br> (9) supporting info. with necessary details <br> (10) cause and effect situations <br> (11) active and passive process of expression <br> (10) Predicting <br> (11) Justifying | (1) Identifying the main trends, features, modalities and tendencies; <br> (2) Identifying similarities, <br> (3) Identifying dissimilarities <br> (4) Inferencing the underlying patterns or movement, or information <br> (5) Predicting future trend, movement, growth, etc. <br> (6) Spotting the underlying reasons, factors, causes, etc. <br> (7) Identifying the technical and logical aspects <br> (8) Identifying cause, effect, significances and impact |

Yamuna Kachru's observations may help decide the source domains of the types of contents, the exercises, etc. the students should be given in the process. Each culture has its own conventions of writing. According to Kachru, it is helpful for the instructors to look at these closely if one is interested to understand the process of writing, especially the cultural context of the product and cultural value assigned to writing. The course learning outcomes of the modules of the writing contents must align with the program learning outcomes. Again, the program learning outcome must align with the education learning outcomes of the country. Kachru's concentric diagram concept [87] might provide a basic framework in this perspective. For examples, students who wish to pursue their career in industrial sector like aviation, oil refinery, or in robotics industries should study specially designed curriculum to develop the skills necessary for the sector. Table 16 shows the tasks, activities, tools, mechanics and cognitive challenges to be given in narrative writing.

**Table 16**. Activities, tools, mechanics, skills and cognitive challenge for narrative writing

| Modules | Activities | Tools, Mechanics & Skills | Cognitive Challenges |
|---|---|---|---|
| Writing Narrative | Narratives, Essays, Critical Appreciations Analytical writing Reviewing Forwarding an argument; opinion; proposition, suggestion, solution, or recommendation, or a crisis, disagreement or uncertainty | Appetizers (Pre-writing activities)<br>Brain-storming, gathering ideas, planning, outlining, organizing, rough-drafting items to put in introduction, body and conclusion)<br><br>Writing the Introduction<br>(1) Paraphrasing the topic/argument and clearly explain your position<br>(2) Using a hook: starting sentence of the introduction<br><br>Writing the Body<br>(1) Description, explanation and illustration with examples (In the body)<br>(2) narrative style (maintaining clarity, brevity and congruence<br>(3) Underlying tone of has to be unbiased, accommodating gender, diversity, and ethics<br>(4) Maintaining a logical sequence in the flow of ideas, information, argument, etc. (In the body)<br>(5) Categorizing, comparing and contrasting,<br>(6) Supporting your argument logically with details,<br>(7) Defending the counter argument,<br>(8) Justifying your position with logical reasoning and evidence-based information<br>(9) Using symbols, metaphors, allusion, ironies, etc. where necessary<br>(10) Generalizing your argument<br>(11) Narrative techniques used in different persons (e.g., first person, third person)<br>Conclusion<br>(12) Rephrasing, inferencing and concluding | (1) Identifying the main argument, purpose, and tone in the proposition or argument;<br>(2) Adopting the correct tone and taking a position and clearly mentioning it; (If you are in dilemma, the essay must focus on both sides of the argument and explain why choosing a side is difficult)<br>(3) Explaining the strong areas of your argument;<br>(4) Explaining your argument from a different perspective other than yours and validating your position<br>(5) Defusing the strong areas of the counter argument, keeping the tone unbiased;<br>(6) Comparing, contrasting and inferencing, when needed;<br>(7) Supporting your argument with logical reasoning and evidence;<br>(8) Using mind map, synonyms and different patterns of expressions in active and passive forms;<br>(9) Brain-storming,<br>(10) Forwarding an opinion, agreement, or disagreement with supporting details and logical reasoning and counter argument |

### 7.1.1.4. Activities, Tools and Cognitive Challenges of Listening Contents and Tasks

The listening recording tracks must reflect diversities in contents that conform to the areas of interest of the leaners in real life. The types of recording tracks should include small talks, dialogues, short and long conversations, monologues, group discussions, podcast, interviews, storytelling, university lectures, public speech, one act plays, description of processes, etc. Recording tracks need to be prepared attribute-wise in CEFR scales. For listening also, the modalities and strengths of different methods and approaches discussed in 6.3.1 – 6.3.5 and 6.4.1 – 6.4.8 should work as the guiding principles in developing the recording tracks of listening. Table 17 shows the types of activities, tools, mechanics, skills and cognitive challenges the cluster domain must have under each listening track.

**Table 17.** Activities, tools, mechanics, skills and cognitive challenges for listening exercises

| Modules | Activities | Tools, Mechanics & Skills | Cognitive Challenges |
|---|---|---|---|
| Listening | Conversation on social and general topics<br>Classroom discussions<br>Lectures<br>Story listening<br>Listening to bulletin<br>Podcast<br>One act play<br>Skit<br>Comics | (1) Taking notes of specific dates, numbers, names, headlines, etc.<br>(2) Listening for factual information<br>(3) Answer transferring<br>(4) Listening for key ideas<br>(5) Separating the supporting details<br>(6) Inferencing<br>(7) Listening for specific information | (1) Identifying speaker's attitudes<br>(2) Identifying speakers' tone<br>(3) Identifying specific information<br>(4) Identifying speaker's purpose<br>(5) Identifying the main ideas<br>(6) Identifying the supporting details |

### 7.1.1.5. Activities, tools and cognitive challenges of Speaking Contents and Tasks

Starting from phrase building to long complex expressions, the speaking activities must also align with preferences of the learners. The level of difficulties of the exercises must match the competence level of the students. Accordingly, a total number of six sets of activities have to be prepared on a single topic in order to accommodate the different competence levels of the students. Again, the modalities and strengths of different methods and approaches discussed in 6.3.1 – 6.3.5 and 6.4.1 – 6.4.8 need to work as the guiding principles in developing the activities for speaking skill. Table 18 shows the activities, tools, mechanics, skills and cognitive challenges for speaking exercises.

**Table 18.** Activities, tools, mechanics, skills and cognitive challenges for speaking exercise.

| Modules | Activities | Tools, Mechanics & Skills | Cognitive Challenges |
|---|---|---|---|
| Speaking | (1) Talking about personal information and two familiar topics<br>(2) Speaking on a topic chosen by the interlocutor for one to two minutes<br>(3) Answering intellectually or cognitively challenging questions on a given familiar topic<br>(4) Story telling<br>(5) talking about everyday life<br>(6) Group discussion<br>(7) Asking questions on everyday affairs<br>(8) Narrating past events<br>(9) Talking about future events, plans, changes, etc. | (1) eloquence<br>(2) fluency<br>(3) accuracy<br>(4) relevance<br>(5) use of vocabulary<br>(6) patterns of expressions<br>(7) perspective - critical overview with analytical explanation<br>(8) diction –<br>(9) pronunciation<br>(10) collocation<br>(11) paraphrasing | (1) Showing fluency, accuracy and soundness in talking about personal likes, dislikes, preference, hobbies, strengths, weakness, family, friends, people you admire, careers, education, etc.<br>(2) Showing proficiency in talking about common and familiar topics<br>(3) Showing insightful understanding while talking about familiar topics<br>(4) Expressing thoughts, opinions, arguments, etc. with reasoning and logical inferences |

### 7.1.2. Goal settings

The course modules, i.e., the contents of reading, writing, listening, and speaking must resonate with Program learning outcomes (PLOs), Course Learning Outcomes (CLOs), Organization Outcomes (OOs), Student's Learning Outcomes (SLOs) and Teacher's Leaning Outcomes (TLOs).

### 7.1.3. Choosing EFL Teaching and Learning approaches for EFL Bigdata Ecosystem

To accommodate the diverse teaching and learning the EFL communities, the study shortlist seven widely used EFL teaching and learning approaches to be used in the cluster domains of the EFL Big Data Ecosystem. The findings of the survey, interviews and qualitative analysis of primary and secondary data substantivate the shortlisting and inclusion of these methods. There methods and approaches are Direct Method, Communicative Language Teaching, Task-Based Learning, Audiolingual Learning, Total Physical Response, Autonomous Learning, and Cognitive Learning.

### 7.1.4. Cognitive Shift

Smart transformation or the DNA of Digital Imprint is a holistic loop involving several critical and technological factors. The stakeholders have to be aware of the potentials of Big Data and Analytics in EFL teaching and learning. It requires management and the head of the institutions to take up the daunting challenges to sustain the cognitive complexities, facilitating the EFL teachers and learners to play their active part in the process of adopting Big Data and Analytics in teaching and learning English. Stakeholders need to understand that Analytics, especially in processing millions of unstructured data, can develop new knowledge, detect insightful patterns, make impactful predictions [168]. It can map out the styles of cognitivism and the learning habits of the students. Further to that, it can take decision faster and better. The stakeholders need to be ready to facilitate AI, humans and machines to learn from each other as part of augmented intelligence. 'Being digital' [169] ,they need to adopt strategies and solutions that will be more reliable, faster and effective, especially with regard to enabling the students to achieve the course learning outcomes and program learning outcomes successfully. Projecting the future development of the students, the EFL teachers need to provide the students with suitable and engaging learning contents. The management needs to arrange robust trainings for the upskilling of the EFL teachers and the students so that they are able to cope up with adaptive learning environment. Such a cognitive shift will enable them to harness the benefits of smart transformation, extending the frontier of teaching and learning beyond the four walls of the classroom. The changes entail a greater responsibility for all the stakeholders in the loop, requiring them to be ready to adapt to constantly shifting power, prospects, potentials and challenges. There are elements of risks and ambiguity, as not that all the data and analytics will lead to discernable patterns and meaningful content development. The institute and the EFL teachers need to remain aware of these risks and ambiguity as well.

### 7.1.5. Policy Formulation

While formulating policy for adaptation of Big Data, the institute needs to evaluate and scrutinize certain crucial factors. First of all, it is important to decide to what extent the institute is going to use Big Data in teaching EFL. Then, the institute needs to identify the potential challenges that can have impactful realities for the stakeholders, especially for the teachers and the learners. Once the ethical issues of protection of privacy are reviewed and decided, the institute needs to focus on how to bring about cognitive changes among the stakeholders so that they can cope up with the smart hype and technological environment. The policies integrating the scopes for harnessing the positive or utopian aspects of the technology should also protect the stakeholders from being exposed to the negative or dystopian aspects of the technology [170]. The policy should algin with the national policy, especially with the guidelines of Ministry of Education, Ministry of Culture and Ministry of Science and Technology. 73% of Chinese think the shift towards AI domains will be positive overall, creating jobs and addressing societal challenges [171]. In fact, formulation of policy to integrate AI in EFL pedagogy will create new employment opportunities for data scientists, data engineers, data consultants, content developers, Video Editor, Content Editor, Networking engineer, and so many, without minimizing opportunities for the EFL teachers in the Institution.

### 7.1.6. Human Capital Development

The focus on developing AI-literate human capital with generic and specific competence should get priority in the formulation of the policy at any level. Having access to so much data doesn't mean that they will always work as the source of actionable material for teaching and learning English. The data and contents to be used for machine learning of the central computer need to have relevance, covering all the skills and subskills of English language. In other words, they must not fall into the category of 'context-less' quantification. To harness the potential of AI, it is important to adequately train the general population in AI [171]. Here the general population refers to EFL teachers and leaners and other member in the organization. The EFL organizations can initially focus on "recruiting, training, promoting, and retaining leading AI workforce" [172] on a smaller scale, who will in the process leverage the EFL teachers in developing new skills in harnessing the ensuing challenges.

### 7.1.7. Collaborative partnership and Charters of New Responsibilities

A big challenge lies in the fact that neither many of these IT professionals are knowledgeable in pedagogical strategies needed to support effective learning, nor the EFL teachers are knowledgeable in IT skills and AI literacy. Therefore, the integration has to be collaborative in nature, both the interest groups supporting each other. It's not just about "doing digital" [169], i.e., not about installing the operating tools and equipment in the institutions like smart boards, uninterrupted wi-fi service, smart devices, access to online materials, resourceful digital library, cloud-based storage facility, and big data tool and software. Rather it is more about "being digital" [169] i.e., getting smatter with one's approach to collect, process, administer, manage, discover, model and distribute unlimited data (learning contents) with regard to facilitating learning into a process of acquisition. All the academic data including the students' assignments, classwork, group works, exam scripts of quizzes, formative tests, progressive tests, and course final exams, presentation videos, audios of speaking skills, drilling sessions, dissertations, project works, etc. can be used as source of raw data for big data analytics to scientifically understand the "knowledge of patterns in data" [79] the students produce. Big data and analytics can also use these patterns in data to cull out trends of cognitivism and learning habits of the students. Interestingly, when the EFL teachers are able to map out the learning habits and styles of cognitivism of the students, they can switch to more effective teaching strategies. They can develop better learning contents and better learning exercises. The jobs of the data engineers will be to use these contents and exercises as structured and semi structured data to teach the machine. In case of moderation of the automatically generated contents and bias check, the data collected from open big data sources can be stored in an intermediary platform before they are stored in the cluster domain of the contents. The more the machine is exposed to huge volume of structured and semi structured data, the more it will be able to detect the variations. Following the insightful findings of data analytics, the institute can also 'set goals' [22] for their students in a scientific way. The job of data engineers is also about preparing evaluation reports of the students, based on their learning habits. If a teacher is aware of the learning habit of the students, he or she can do the necessary changes in the process of teaching, making the approach more scientific and goal oriented. The data engineers also need to maintain, store and run the dynamic platform in institution's cloud-based storage, and ensure its e effectiveness. The task of the teacher at the initial phase is to develop structured and semi-structured data for populating the cluster domain of the contents.

### 7.1.8. Logistics

With regard to knowledge building about logistics, the institute needs to focus on machine learning, Information Technological support [61], Internet connectivity [61], Learning Management Systems (LMS) blackboard [173], database server, maintenance and manpower support [61]. During the COVID19 pandemic period, the education system of the world had undergone a massive paradigm shift from onsite to online mode of teaching and learning. Many institutions have started using LMS blackboard platforms for their learners, installing strong broadband internet connectivity for their faculty members and administrators. The trend is still ongoing. The teaching and learning communities are already familiar with the ways technologies are adapted in the dissemination of academic knowledge. The study finds that many universities in the Kingdom of Saudi Arabia [173] and Japan already have the basic logistic supports for the implementation of the smart transformation, provided they should develop their EFL bigdata ecosystem Teachers and students are digitally and centrally connected by LMS Blackboard which carry the digital imprints of all pros and cons of the courses in every semester [173]. The campuses have strong and uninterrupted broadband connection. Classrooms have smartboards [173]. Teachers and students have smart devices like laptops, note pads and smart phones. Microsoft offers its corporate client services to all the faculty members and staffs. The only thing needed is to tune up entire systems from 'doing digital' to 'being digital' [140]. In other words, the stakeholders need to get smarter in their approach to collect, process, administer, manage, discover, model and distribute unlimited data (learning contents) and strive to facilitate the process of learning into a process of acquisition.

### 7.1.9. Pragmatic

In pragmatic phase the stakeholders need to gather adequate knowledge on collaborative partnership building between data scientists and EFL teachers and learners. They need to develop the tools    to integrate, assess, monitor and evaluate the performance of the EFL teachers and learners. They need to use the predictive analytics for the improvement of the performances of the EFL teachers and learners. They need to develop a data interoperability model. They need to have framework for systems operation, streaming of data, monitoring and upgradation. Having an integrated framework to cover all these issues is important for successful implementation of EFL Bigdata ecosystem.

## 7.2. Machine Learning and Algorithms

The goal of this work is to leverage machine learning (ML) techniques to enhance the learning capability by providing personalized course content based on individual student's skill level, preferred learning style, and their interests in specific subjects. The goal also includes individual preference of the teachers with regard to choosing methods of teaching. For instance, students with Reading Knowledge Zone as RA1-DM-AC indicates that they possess a very basic reading ability, respond well to direct teaching methods, and have a strong preference for arts and culture.  And the teacher teaching them, based on these preferences, also decides to follow direct method for them. In this case, the machine learning system will tailor the contents to provide the student with introductory reading materials related to arts and culture. This personalized approach aims to cater to the unique needs and preferences of each student. The ultimate goal is to optimize the learning experience by leveraging machine learning to create tailored content that aligns with the divers teaching and learning needs of the EFL communities. The machine learning system will be trained using a carefully curated training dataset that consists of various course contents labeled with appropriate indicators. These labels will enable the machine to identify the skill level (basic, intermediate, or advanced) of each piece of content. Additionally, the dataset will include classifications for different subjects, such as Arts and Culture, Technical, Current Affairs, and others. During the training process, the machine learning model will learn from the labeled dataset, analyzing the features and patterns within the course contents. By examining the characteristics and indicators associated with each content item, the model will develop the ability to classify them according to skill levels, teaching modalities and subjects accurately. Eventually system will recommend contents accordingly through iterative training and adjustments. There are several machine learning algorithms that could be suitable for the task of personalized course content recommendation

**Decision Trees**: Decision trees are versatile algorithms that can handle both classification and regression tasks. They can be effective for determining skill levels, teaching modalities, learning styles, and subject classifications based on the provided indicators [174].

**Random Forests**: Random forests are an ensemble learning method that combines multiple decision trees. They can handle complex feature interactions and provide robust predictions. Random forests could be useful for classifying course content based on skill levels, teaching modalities and subjects [175].

**Support Vector Machines (SVM):** SVMs are powerful algorithms for classification tasks. They can handle high-dimensional data and are effective at separating different classes. SVMs could be employed to classify course content based on skill levels and subjects [176].

**Naive Bayes:** Naive Bayes is a probabilistic classifier that works well with text classification tasks. It could be utilized to classify course content into different subjects based on text features or indicators [177].

**K-Nearest Neighbors (KNN):** KNN is a simple and intuitive algorithm that classifies new data based on its proximity to labeled data points. KNN could be applied to classify course content based on similarity to labeled materials [178].

**Neural Networks:** Deep learning models, such as neural networks, have shown great success in various classification tasks. They can learn complex patterns and relationships in the data. Neural networks could be trained to classify course content based on skill levels, learning styles, and subject preferences [179].

According to Dr. Mahmudur Rashid, Senior Data and Infrastructure Engineer at Schlumberger, "The Data scientists need to experiment with multiple algorithms and evaluate their performance based on the standard metric (F1 Score, Accuracy, Confusion matrix etc.) to determine the most suitable approach for this specific task."

### 7.2.1. Machine Learning Cluster Domain for EFL Bigdata Ecosystem

The machine needs to be begiven inputs [157] by the data scientists in the form of cluster domains developed by the content developers or EFL teachers. The relevant characteristics of the data hosted under each cluster domain are pre-defined following the modalities and manuals for content development of different teaching and learning approaches including DM, TPR, CLT, TBL, CL, ATL and CL. How to develop them have already been discussed in previous section. As we know that these cluster-domains should be skilled-wise (reading, writing, listening, and speaking) and attribute-wise according to the six attributes of CEFR scales (A1, A2, B1, B2, C1 & C2). Their difficulty levels shall match with the respective attributes. For examples, if the users click on the hypertext like RA1, it will direct the users to contents that refer to the reading content suitable for the beginners. Accordingly, WA2, LA1 and SA1 to writing, listening and speaking contents that are suitable for the beginners. Likewise, RB1, WB1, LB1, and LB1 will direct to reading, writing, listening and speaking contents that are suitable for the intermediate level students. Finally, RC1, WC1, LC1, and SC1 will direct to the reading, writing, listening and speaking contents that are suitable for advanced level students. It is important to note that there are differences between the contents stored under RA1 and RA2. The contents stored in RA2 is one stage ahead of the contents that are stored in RA1. There will be, in fact, 24 different types of hypertexts referring to 24 different cluster domains for the four skills.

The machine can be taught to recognize the underline patterns in the contents of each cluster domain. In the beginning it needs to be taught to recognize different parts of speech. Then, with the help of signs the functioning mechanics of English expressions can be given a sing-centric expressions. Next, the machine needs to be taught to recognize the common patterns underlining in short and long phrases, and in simple, complex and compound expressions. In fact, signs can be used to replicate the underlying patterns in English expressions [180]. Figure 4 shows the use of signs to replicate the words in their parts of speech form.

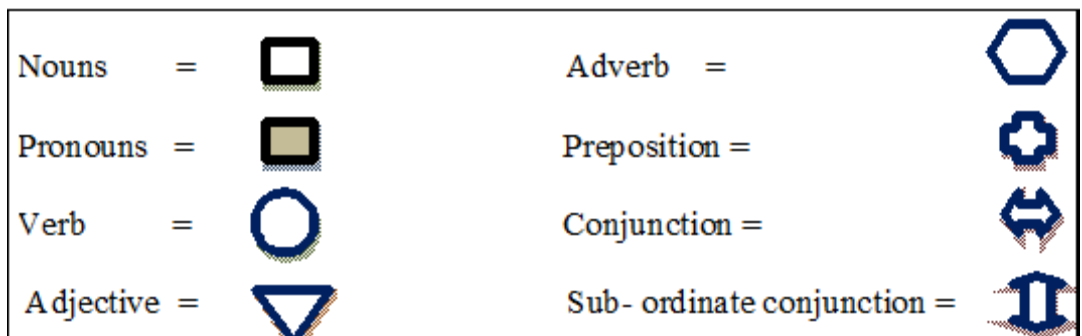

**Figure 4**. Use of signs to recognize the patters of expression

The machine can recognize the patterns of expressions in English sentences, regardless their structure, size and complexities. However, before that it needs to be populated with the common patterns existing in English language. Image 5 shows a few most widely used patterns in use in English. A few common patters of phrases, sentences and questions are as followings and all these patterns can be transformed in to sign language to teach the machine.

*preposition + noun (at hotel)*
*adjective + noun      (good man)          .*
*article + adjective + noun       (a fast car)*
*article + adverb+ adjective + noun (a beautifully maintained house)*
*preposition + article + adjective + noun (in the green field)*
*noun + noun          (English Teacher)*
*subject + verb + noun. (Water saves life)*
*subject + verb + adjective (Helicopter flies high)*
*subject + verb + adverb (Tram moves slowly.)*
*subject + verb + preposition + noun (Tress grow in forest.)*
*subject + verb. (Bird flies.)*
*Subject + verb + conjunction + verb + noun (He works and earns money.)*
*WH + Helping verb + subject + verb + preposition+ article + noun? (Where does she live in the city?)*
*WH + helping verb + subject + verb + adverb? (Why do they work there?)*
*Subordinate conjunction + subject + verb + subject + verb.  (If water is heated, it evaporates.)*

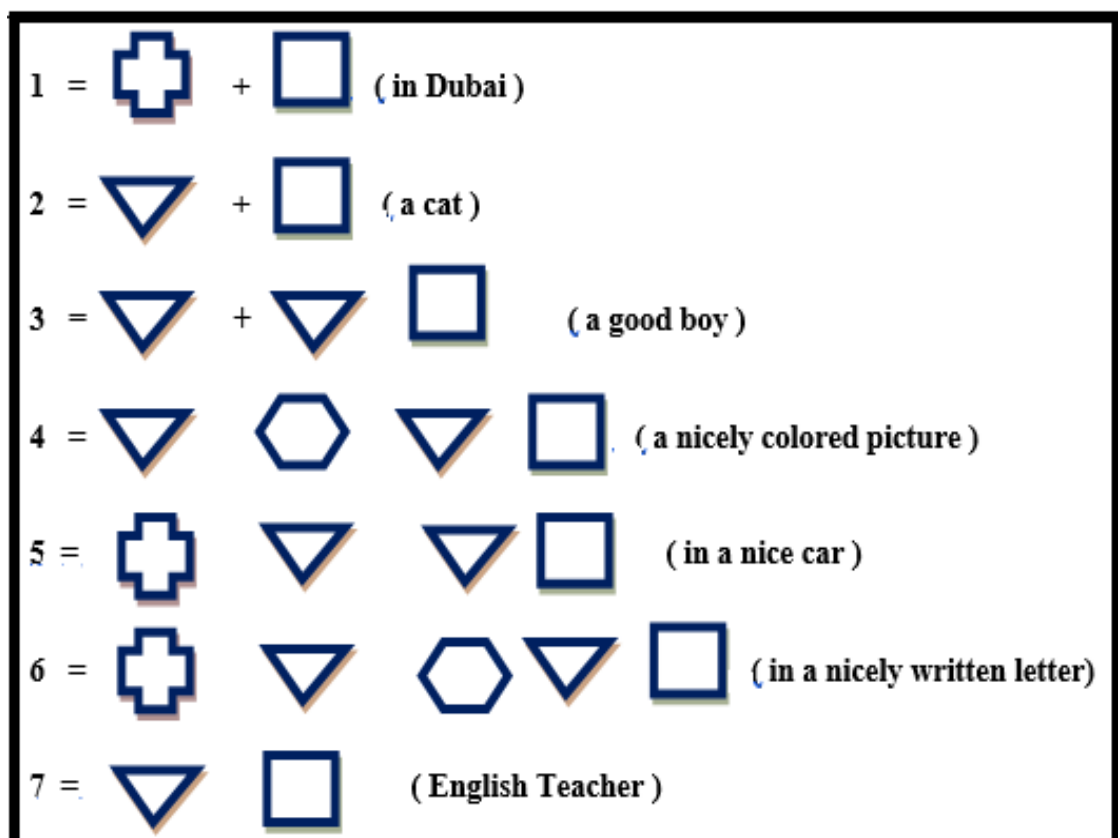

**Figure 5.** Common patterns of expressions in English in Sign Language

Using the same signs, machine can be taught to recognize the incomplete expressions as well. Figure 6 shows the ways to teach machine to recognize the patterns underlying in incomplete expressions.

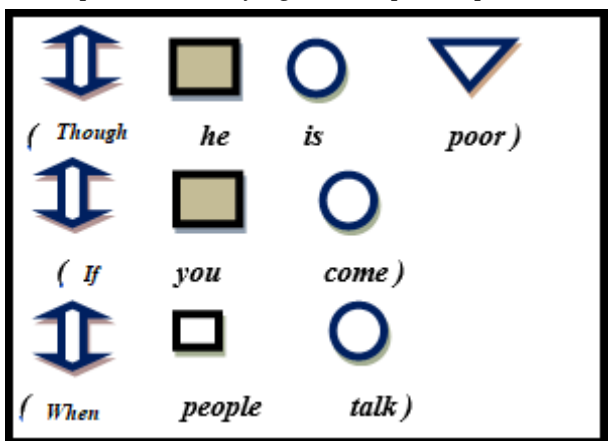

**Figure 6.** The patterns underlying in incomplete expressions in English

In the same way, the machine can be taught to recognize the patterns underlying in different sentences. Figure 7 shows the underlying patterns in complete expressions.

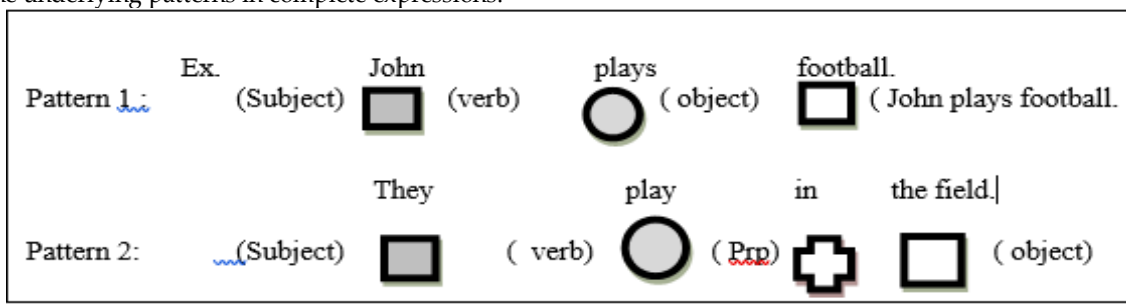

**Figure 7.** The patterns underlying in complete expressions.

The objective of creating these cluster domains with supervised and structured data is to teach the machine to identity the patterns that exist in the contents of each of these cluster domains. When machine will be taught to detect the patterns existing in each of these cluster domains, it can identify similar patterns in the unstructured data pools of Big Data without being supervised [159]. Over the period of time, when each of these cluster domains will be populated with huge volume of structured and semi-structured data, it will become more efficient in extracting similar data from open sources. Data scientists can try algorithms such **as Decision** Trees, Random Forecasts, Support Vector Machines and Naïve Bayes for these.

**Table 19.** Hyperlink for Cluster Domains for Skill-wise contents

| | Attributes (Parameters) in CEFR Scale | | | | | |
|---|---|---|---|---|---|---|
| | Beginner | | Intermediate | | Advanced | |
| **Skills** | **A1** | **A2** | **B1** | **B2** | **C1** | **C2** |
| **Reading** | RA1 | RA2 | RB1 | RB2 | RC1 | RC2 |
| **Writing** | WA1 | WA2 | WB1 | WB2 | WC1 | WC2 |
| **Listening** | LA1 | LA2 | LB1 | LB2 | LC2 | LC2 |
| **Speaking** | SA1 | SA2 | SB1 | SB2 | SC1 | SC2 |

### 7.3. Integration Phase

The cluster domains of contents need to be integrated with the chosen EFL teaching and learning approaches. Integration phase refers to machine learning [156] for aligning the Cluster Domain of Contents with EFL teaching and learning approaches attribute-wise. So, the new hypertext for reading, for examples, will be like RA1-DM, refereeing to the reading contents that are prepared following the modalities, manuals and activities of Direct Method. Likewise, WC2-TBL refers to writing contents suitable for task-based method. SB2-CLT refers to speaking contents suitable for communicative language teaching and the like. Integration of the 24 hypertexts for the cluster domains with hypertexts of seven different EFL teaching and learning approaches will produce 168 unique hypertexts of cluster domains, each directing the users to different types of practice materials prepared following specific modalities on the same topic. In other words, one little reading passage on 'Entertainment' will have 42 different types of contents, each one being prepared method-wise and attribute-wise to suite its difficulty level and modality requirement. The algorithms to be tried are mainly but not limited to Decision Trees [174], Random Forests [175], Support Vector Machines, Naïve Bayes, and KNN. The data scientists need to try them and find the suitable ones based on their accuracy in results  Table 20, 21,22,23,24, and 25 show how the hyperlinks for mapping the cluster domain (skill-wise and attribute-wise) contents with the EFL teaching and learning approaches.

**Table 20.** Tool for mapping teaching and learning needs of the teachers and learners

| Attribute | | Resonating Teaching and Learning Approaches | | | | | | |
|---|---|---|---|---|---|---|---|---|
| AI | | DM | CLT | TBL | AL | TPR | ATL | CL |
| Skills | Reading | RA1-DM | RA1-CLT | RA1-TBL | RA1-AL | RA1-TPR | RA1-ATL | RA1-CL |
| | Writing | WA1-DM | WA1-CLT | WA1-TBL | WA1-AL | WA1-TPR | WA1-ALT | WA1-CL |
| | Listening | LA1-DM | LA1-CLT | LA1-TBL | LA1-AL | LA1-TPR | LA1-ALT | LA1-CL |
| | Speaking | SA1-DM | SA1-CLT | SA1-TBL | SA1-AL | SA1_TPR | SA1-ALT | SA1-CL |
| | TPR* | DM* Direct Method; CLT* Communicative Language Teaching; | | | | | | |
| CEFR | Total | TBL* Task-Based Learning; AL* Audiolingual Learning; | | | | | | |
| Beginner | Physical Response | ATL*Autonomous Learning; CL* Cognitive Learning | | | | | | |

**Table 21.** Tool for mapping teaching and learning needs of the teachers and learners

| Attribute | | Resonating Teaching and Learning Approaches | | | | | | |
|---|---|---|---|---|---|---|---|---|
| A2 | | DM | CLT | TBL | AL | TPR | ATL | CL |
| Skills | Reading | RA2-DM | RA2-CLT | RA2-TBL | RA2-AL | RA2-TPR | RA2-ATL | RA2-CL |
| | Writing | WA2-DM | WA2-CLT | WA2-TBL | WA2-AL | WA2-TPR | WA2-ALT | WA2-CL |
| | Listening | LA2-DM | LA2-CLT | LA2-TBL | LA2-AL | LA2-TPR | LA2-ALT | LA2-CL |
| | Speaking | SA2-DM | SA2-CLT | SA2-TBL | SA2-AL | SA2_TPR | SA2-ALT | SA2-CL |
| | TPR* | DM* Direct Method; CLT* Communicative Language Teaching; | | | | | | |
| CEFR | Total | TBL* Task-Based Learning; AL* Audiolingual Learning; | | | | | | |
| Beginner | Physical Response | ATL*Autonomous Learning; CL* Cognitive Learning | | | | | | |

**Table 22.** Tool for mapping teaching and learning needs of the teachers and learners

| Attribute | | Resonating Teaching and Learning Approaches | | | | | | |
|---|---|---|---|---|---|---|---|---|
| B1 | | DM | CLT | TBL | AL | TPR | ATL | CL |
| Skills | Reading | RB1-DM | RB1-CLT | RB1-TBL | RB1-AL | RB1-TPR | RB1-ATL | RB1-CL |
| | Writing | WB1-DM | WB1-CLT | WB1-TBL | WB1-AL | WB1-TPR | WB1-ALT | WB1-CL |
| | Listening | LB1-DM | LB1-CLT | LB1-TBL | LB1-AL | LB1-TPR | LB1-ALT | LB1-CL |
| | Speaking | SB1-DM | SB1-CLT | SB1-TBL | SB1-AL | SB1_TPR | SB1-ALT | SB1-CL |
| | TPR* | DM* Direct Method; CLT* Communicative Language Teaching; | | | | | | |
| CEFR | Total | TBL* Task-Based Learning; AL* Audiolingual Learning; | | | | | | |
| Intermediate | Physical Response | ATL*Autonomous Learning; CL* Cognitive Learning | | | | | | |

**Table 23.** Tool for mapping teaching and learning needs of the teachers and learners

| Attribute | | Resonating Teaching and Learning Approaches | | | | | | |
|---|---|---|---|---|---|---|---|---|
| B2 | | DM | CLT | TBL | AL | TPR | ATL | CL |
| Skills | Reading | RB2-DM | RB2-CLT | RB2-TBL | RB2-AL | RB2-TPR | RB2-ATL | RB2-CL |
| | Writing | WB2-DM | WB2-CLT | WB2-TBL | WB2-AL | WB2-TPR | WB2-ALT | WB2-CL |
| | Listening | LB2-DM | LB2-CLT | LB2-TBL | LB2-AL | LB2-TPR | LB2-ALT | LB2-CL |
| | Speaking | SB2-DM | SB2-CLT | SB2-TBL | SB2-AL | SB2_TPR | SB2-ALT | SB2-CL |
| | TPR* | DM* Direct Method; CLT* Communicative Language Teaching; | | | | | | |
| CEFR | Total | TBL* Task-Based Learning; AL* Audiolingual Learning; | | | | | | |
| Intermediate | Physical Response | ATL*Autonomous Learning; CL* Cognitive Learning | | | | | | |

**Table 24.** Tool for mapping teaching and learning needs of the teachers and learners

| Attribute | | Resonating Teaching and Learning Approaches | | | | | | |
|---|---|---|---|---|---|---|---|---|
| C1 | | DM | CLT | TBL | AL | TPR | ATL | CL |
| Skills | Reading | RC1-DM | RC1-CLT | RC1-TBL | RC1-AL | RC1-TPR | RC1-ATL | RC1-CL |
| | Writing | WC1-DM | WC1-CLT | WC1-TBL | WC1-AL | WC1-TPR | WC1-ALT | WC1-CL |
| | Listening | LC1-DM | LC1-CLT | LC1-TBL | LC1-AL | LC1-TPR | LC1-ALT | LC1-CL |
| | Speaking | SC1-DM | SC1-CLT | SC1-TBL | SC1-AL | SC1_TPR | SC1-ALT | SC1-CL |
| | TPR* | DM* Direct Method; CLT* Communicative Language Teaching; | | | | | | |
| CEFR | Total | TBL* Task-Based Learning; AL* Audiolingual Learning; | | | | | | |
| Advance | Physical Response | ATL*Autonomous Learning; CL* Cognitive Learning | | | | | | |

**Table 25.** Tool for mapping teaching and learning needs of the teachers and learners

| Attribute | | Resonating Teaching and Learning Approaches | | | | | | |
|---|---|---|---|---|---|---|---|---|
| C2 | | DM | CLT | TBL | AL | TPR | ATL | CL |
| | | | | | | | | |
| Skills | Reading | RC2-DM | RC2-CLT | RC2-TBL | RC2-AL | RC2-TPR | RC2-ATL | RC2-CL |
| | Writing | WC2-DM | WC2-CLT | WC2-TBL | WC2-AL | WC2-TPR | WC2-ALT | WC2-CL |
| | Listening | LC2-DM | LC2-CLT | LC2-TBL | LC2-AL | LC2-TPR | LC2-ALT | LC2-CL |
| | Speaking | SC2-DM | SC2-CLT | SC2-TBL | SC2-AL | SC2_TPR | SC2-ALT | SC2-CL |
| | TPR* | DM* Direct Method; CLT* Communicative Language Teaching; | | | | | | |
| CEFR | Total | TBL* Task-Based Learning; AL* Audiolingual Learning; | | | | | | |
| Advance | Physical Response | | ATL*Autonomous Learning; CL* Cognitive Learning | | | | | |

## 7.4. Personalized Preferences

Machine can be taught to categorize the areas of interest of the EFL learners to create a personalized domains for learning experience. Data scientists can apply algorithms such as Support Vector Machines for classification of the contents. While categorizing, the types of terminologies to be used need to be observed carefully. Because, 'terminology matters [144]. Usually, they are very slippery with pitfalls. So, while grouping or categorizing the areas of interest, the content writers or data scientists or the teacher working as a domain consultant in their new role, should select a range for each group. They should ensure that the tone is neutral, not aggressive or discriminatory. The guiding principle is that the learners should exert a sense of familiarity [36] [154] in their learning environment [127]. The developers/teachers can use their own ways, if they are found more convenient. The following is the list of Categories of Areas of Internets of the EFL learners, prepared on the basis of the findings of literature review [153]. Each of these hypertexts will direct to the contents that reflect the category it refers to. For example, AI will direct to category of contents that cover agriculture and industry, FB to fashion and beauty, NM to newspaper and media, SM to social media, etc.  Table 26 shows hypertext for categories of personalized preferences of the learners for different areas or subjects of interests.

**Table 26.** Hypertexts of the areas of the interest of the EFL learners

| Category | Hypertext | Category | Hypertext | Category | Hypertext |
|---|---|---|---|---|---|
| Agriculture and Industries | AI | Technology & Research | TR | Science & Intelligence | SI |
| Art & Culture | AC | Newspapers & Media | NM | Social Media | SM |
| Fashion & Beauty | FB | Politics & Leaders | PL | Law & Government | LG |
| Travel & Adventure | TA | History & Tradition | HT | Faith & Values | FV |
| Music & Performers | MP | Outdoors & Hobbies | OH | Gaming & Youth | GY |
| Careers & Skills | CS | Comics & Animations | CA | Literature & Society | LC |

## 7.5. Resonating

Personalized experience in EFL teaching and learning is possible when teachers can choose the approaches they prefer to use in teaching and learners can choose the learning contents they prefer to read, listen, speak and write about. At this stage, the machine learning [160] resonates the skill-wise and attribute-wise cluster domains of contents with the personalized areas of interests of the EFL learners. Data scientist can apply multiple algorithms such as Decision Tress, Random Forests, Naïve Bayes, SVM, etc. and decide the best ones on the basis of the accuracy of results in resonating. Table 27 shows the ways to resonate the personalized preferences of the teachers for methods with the existing linguistic competence of the learners.

**Table 27.**  Hypertexts of Mapping the teacher's preferences with existing linguistic potential of the learners

| Students | Skills and Teaching and Learning Approaches | | | |
|---|---|---|---|---|
| | Reading | Writing | Listening | Speaking |
| AA1 | RA1-DM | WA1-TBL | LA1-AL | SA1-CLT |
| AA2 | RA2-DM | WA1-TBL | LA1-AL | SA1-TPR |
| AA3 | RA1-DM | WA1-CLT | LA12-AL | SA1-DM |
| AA4 | RA1-DM | WA1-TBL | LA1-AL | SA1-CLT |
| AA5 | RA2-CLT | WA1-TBL | LA1-DM | SA1-AL |
| AA6 | RA1-DM | WA1-CLT | LA2-AL | SA1-DM |
| AA7 | RB1-CLT | WB1-TBL | LB1-AL | SA2-CLT |

| AA8 | RA2-DM | WA1-TBL | LA1-AL | SA1-CLT |
| AA9 | RA1-DM | WA1-TBL | LA1-AL | SA1-CLT |
| BB1 | RA1-DM | WA1-TBL | LA1-AL | SA1-CLT |
| BB2 | RB2-ATL | WA2-CLT | LB2-AL | SA2-CLT |
| BB3 | RA1-DM | WA1-CLT | LA12-AL | SA1-DM |
| BB4 | RA1-DM | WA1-TBL | LA1-AL | SA1-CLT |
| BB5 | RB1-CLT | WB1-DM | LB1-AL | SA2-CLT |
| BB6 | RA2-TBL | WA1-CLT | LA1-AL | SA1-TBL |
| BB7 | RA1-DM | WA1-CLT | LA2-AL | SA1-DM |
| BB8 | RA1-DM | WA1-TBL | LA1-AL | SA1-CLT |
| BB9 | RA1-DM | WA1-TBL | LA1-AL | SA1-CLT |
| CC1 | RB1-TBL | WB1-DM | LB1-AL | SA2-CLT |
| CC2 | RA1-DM | WA1-TBL | LA1-AL | SA1-CLT |

At first, a teacher having a class of 20 students has been interviewed to know the types of teaching approaches he prefers for his students. The teacher makes an assessment of the linguistic ability of the students. He identifies the strengths and weaknesses of the students and then he chooses the method he prefers for each student. Table 14. below shows the details of the survey. From AA1 to CC2 includes 20 students. Their academic potentials in different language skills are mentioned to right columns, following their ID numbers. Just immediately after the hypertext reflecting their competence status, the method that is preferred for them by the teacher is mentioned. For example, RA2-DM after AA1 means that the student carrying the ID number AA1 language potential in reading matches the competence level of A1 in CEFR scale and his teacher prefers to teach his reading skill following direct method. As the academic potentials of the students are different in different skills, the teacher prefers to use different methods for different learners. For examples, students bearing the ID number AA5 has different competence level in different skills. His reading skill competence corresponds to A2 level in CEFR scale and his teacher prefers for him CLT method to teach him reading skill. However, his writing competence attributes to A1 in CEFR scale and his teacher prefers to use TBL for improving his writing skill. Again, his Listening and Speaking competence corresponds to A1 level and the teacher chooses DM for teaching him listening and Al for speaking skill.

Next, the teacher carries out a survey among the EFL learners to find out the areas that they really enjoy. He gives four choices to the students. The students choose four areas that they personally love to learn about. . They are told to use 1, 2, 3, 4 under the box that best represent the topics or subjects or fields that engage them passionately. The following chart (Table 28) provides the ways this data has to be collected to identify the peronalized areas of interest of the EFL learners.

**Table 28.** Tool for mapping the areas of the interest of the EFL learners

| Stude nts | Categories of Areas of Interest | | | | | | | | | | | | | | | | | | | | | | | |
|---|---|---|---|---|---|---|---|---|---|---|---|---|---|---|---|---|---|---|---|---|---|---|---|---|
| | A I | A C | F B | T A | M P | C S | F H | M H | T R | N M | P L | H T | O H | C A | F H | E C | G Y | L C | F R | S E | S I | S M | L G | F V |
| AA1 | | 1 | | 2 | | | | 3 | | | | | | | | | 4 | | | | | | | |
| AA2 | | | | | | 1 | | 2 | | | | 3 | | | | | 4 | | | | | | | |
| AA3 | | | | | 4 | | 2 | | | | | | | | | | 1 | | | | | | 3 | |
| AA4 | | | | 1 | | | | 2 | | | | | | 3 | | | | | 4 | | | | | |
| AA5 | | | 3 | | | | 1 | | | | 2 | | | | | 4 | | | | | | | | |
| AA6 | | | | | | | | | 3 | | | | | | | 2 | | | | 4 | 1 | | | |
| AA7 | | 4 | | 3 | | 2 | | | | | | | | | | | | | | 1 | | | | |
| AA8 | 4 | | | | | | | 3 | | | | | | 2 | | | | | | 1 | | | | |
| AA9 | | | | | | 4 | | | 2 | 3 | | | | | | | | | | 1 | | | | |
| BB1 | | | | | 4 | | | | 3 | 2 | | | | | | | | | | 1 | | | | |
| BB2 | | 4 | | | | 3 | 1 | | 2 | | | | | | | | | | | | | | | |
| BB3 | | | | 4 | | | | 2 | | | | | | | | | | | | 1 | | | 3 | |
| BB4 | | | | 1 | | | | 3 | | | | | | 4 | | | | | | 3 | | | | |
| BB5 | | | 3 | | | | | 4 | | | | | | 1 | | | | | | 2 | | | | |
| BB6 | | | 1 | | | 2 | | 4 | | | | | | | | | | | 3 | | | | | |
| BB7 | | | | | | | | 1 | | 4 | | | | 3 | | | | | | 2 | | | | |
| BB8 | | | | 1 | | 3 | | | | | | | | 4 | | 2 | | | | | | | | |
| BB9 | | | | | | 4 | | 3 | | | | | | 1 | | | | | | 2 | | | | |
| CC1 | | | | 1 | | | | 3 | | | | | | 4 | | | | | | 2 | | | | |
| CC2 | | | | 1 | | | | 4 | | | | 3 | | | | | | | | 2 | | | | |

Finally, the personalized preference of the teacher is resonated with the personalized preferences of the students, making the entire approach absolutely personalized and engaging. The teacher has chosen the method he preferred for a student, and the student has chosen the contents that engages him passionately. The following hyperlink (Table 29) shows the resonation of teaching approaches with the types of learning contents the students love.

**Table 29.** Mapping the teacher's preferences with existing linguistic potential of the learners

| Students | Skills, Teaching and Learning Approaches, and Interest Areas | | | |
|---|---|---|---|---|
| | Reading | Writing | Listening | Speaking |
| AA1 | RA1-DM-AC | WA1-TBL-TA | LA1-AL-MH | SA1-CLT-GY |
| AA2 | RA2-DM-CS | WA1-TBL-TR | LA1-AL-HT | SA1-TPR-GY |
| AA3 | RA1-DM-GY | WA1-CLT-FH | LA1-AL-HT | SA1-DM-MP |
| AA4 | RA1-DM-TA | WA1-TBL-TR | LA1-AL-FH | SA1-CLT-FR |
| AA5 | RA2-CLT-FB | WA1-TBL-MH | LA1-DM-HT | SA1-AL-EC |
| AA6 | RA1-DM-PL | WA1-CLT-GY | LA1-AL-SM | SA1-DM-SE |
| AA7 | RB1-CLT-AC | WB1-TBL-TA | LB1-AL-FH | SA2-CLT-SE |
| AA8 | RA2-DM-CA | WA1-TBL-SE | LA1-AL-AI | SA1-CLT-FH |
| AA9 | RA1-DM-TR | WA1-TBL-CS | LA1-AL-NM | SA1-CLT--SE |
| BB1 | RA1-DM--MP | WA1-TBL-HT | LA1-AL-SE | SA1-CLT-PL |
| BB2 | RB2-ATL-FH | WA2-CLT-MP | LB2-AL-ATL | SA2-CLT-PL |
| BB3 | RA1-DM-SE | WA1-CLT-TA | LA1-AL-TR | SA1-DM-LG |
| BB4 | RA1-DM-TA | WA1-TBL-FH | LA1-AL-TR | SA1-CLT-SE |
| BB5 | RB1-CLT-TA | WB1-DM-TR | LB1-AL-CA | SA2-CLT-SE |
| BB6 | RA2-TBL-TA | WA1-CLT-MH | LA1-AL-TR | SA1-TBL-FR |
| BB7 | RA1-DM-TR | WA1-CLT-PL | LA1-AL-FH | SA1-DM-SE |
| BB8 | RA1-DM-FH | WA1-TBL-TA | LA1-AL-CA | SA1-CLT-GY |
| BB9 | RA1-DM-TR | WA1-TBL-CS | LA1-AL-CA | SA1-CLT-SE |
| CC1 | RB1-TBL-TR | WB1-AL-TA | LB1-AL-SE | SA2-CLT-CA |
| CC2 | RA1-DM-OH | WA1-TBL-TA | LA1-AL-TR | SA1-CLT-SE |

For examples, RA2-DM-AC means the student bearing the ID number AA1 enjoys reading about Art and Culture as his number 1 (see Table 28 and Table 29) preference and his existing knowledge potential in reading skill corresponds to beginner level (A1) in CEFR scale, and the teacher would prefer to adapt the contents, tasks and activities that conform to the modalities of Direct Method for this student to teach him reading skill.

### 7.6. Zone of Proximity Development

The framework for developing the zone of proximity development is based on the widely accepted theories of language acquisition and input hypothesis propounded by Krashen and Terrel. Table 30 shows the proximity development framework.

**Table 30.** Zone of Proximity Development

| Student ID | Zone of Proximal Development | | Zone of Proximal Development | | Zone of Proximal Development | | Zone of Proximal Development | |
|---|---|---|---|---|---|---|---|---|
| | Current Knowledge Zone Reading | Potential Knowledge Zone Reading | Current Knowledge Zone Writing | Potential Knowledge Zone Writing | Current Knowledge Zone Listening | Potential Knowledge Zone Listening | Current Knowledge Zone Speaking | Potential Knowledge Zone Speaking |
| AA1 | RA1-DM-AC | RA2-DM-AC | WA1-TBL-TA | WA2-TBL-TA | LA1-AL-MH | LA2-AL-MH | SA1-CLT-GY | SA2-CLT-GY |
| AA2 | RA2-DM-CS | RB1-DM-CS | WA1-TBL-TR | WA2-TBL-TR | LA1-AL-HT | LA2-AL-HT | SA1-TPR-GY | SA2-TPR-GY |
| AA3 | RA1-DM-GY | RA2-DM-GY | WA1-CLT-FH | WA2-CLT-FH | LA1-AL-HT | LA2-AL-HT | SA1-DM-MP | SA2-DM-MP |
| AA4 | RA1-DM-TA | RA2-DM-TA | WA1-TBL-TR | WA2-TBL-TR | LA1-AL-FH | LA2-AL-FH | SA1-CLT-FR | SA2-CLT-FR |
| AA5 | RA2-CLT-FB | RB1-CLT-FB | WA1-TBL-MH | WA2-TBL-MH | LA1-DM-HT | LA2-DM-HT | SA1-AL-EC | SA2-AL-EC |
| AA6 | RA1-DM-PL | RA2-DM-PL | WA1-CLT-GY | WA2-CLT-GY | LA1-AL-SM | LA2-AL-SM | SA1-DM-SE | SA2-DM-SE |
| AA7 | RB1-CLT-AC | RB2-CLT-AC | WB1-TBL-TA | WB2-TBL-TA | LB1-AL-FH | LB2-AL-FH | SA2-CLT-SE | SB1-CLT-SE |
| AA8 | RA2-DM-CA | RB1-DM-CA | WA1-TBL-SE | WA2-TBL-SE | LA1-AL-AI | LA2-AL-AI | SA1-CLT-FH | SA2-CLT-FH |
| AA9 | RA1-DM-TR | RA2-DM-TR | WA1-TBL-CS | WA2-TBL-CS | LA1-AL-NM | LA2-AL-NM | SA1-CLT--SE | SA2-CLT--SE |
| BB1 | RA1-DM--MP | RA2-DM--MP | WA1-TBL-HT | WA2-TBL-HT | LA1-AL-SE | LA2-AL-SE | SA1-CLT-PL | SA2-CLT-PL |
| BB2 | RB2-ATL-FH | RC1-ATL-FH | WA2-CLT-MP | WA2-CLT-MP | LB2-ATL-AC | LC1-ATL-AC | SA2-CLT-PL | SB1-CLT-PL |
| BB3 | RA1-DM-SE | RA2-DM-SE | WA1-CLT-TA | WA2-CLT-TA | LA1-AL-TR | LA2-AL-TR | SA1-DM-LG | SA2-DM-LG |
| BB4 | RA1-DM-TA | RA2-DM-TA | WA1-TBL-FH | WA2-TBL-FH | LA1-AL-TR | LA2-AL-TR | SA1-CLT-SE | SA2-CLT-SE |
| BB5 | RB1-CLT-TA | RB2-CLT-TA | WB1-DM-TR | WB2-DM-TR | LB1-AL-CA | LB2-AL-CA | SA2-CLT-SE | SB1-CLT-SE |
| BB6 | RA2-TBL-TA | RB1-TBL-TA | WA1-CLT-MH | WA2-CLT-MH | LA1-AL-TR | LA2-AL-TR | SA1-TBL-FR | SA2-TBL-FR |

| BB7 | RA1-DM-TR | RA2-DM-TR | WA1-CLT-PL | WA2-CLT-PL | LA1-AL-FH | LA2-AL-FH | SA1-DM-SE | SA2-DM-SE |
| BB8 | RA1-DM-FH | RA2-DM-FH | WA1-TBL-TA | WA2-TBL-TA | LA1-AL-CA | LA2-AL-CA | SA1-CLT-GY | SA2-CLT-GY |
| BB9 | RA1-DM-TR | RA2-DM-TR | WA1-TBL-CS | WA2-TBL-CS | LA1-AL-CA | LA2-AL-CA | SA1-CLT-SE | SA2-CLT-SE |
| CC1 | RB1-TBL-TR | RB2-TBL-TR | WB1-TBL-TA | WB2-TBL-TA | LB1-AL-SE | LB2-AL-SE | SA2-CLT-CA | SB1-CLT-CA |
| CC2 | RA1-DM-OH | RA2-DM-OH | WA1-TBL-TA | WA2-TBL-TA | LA1-AL-TR | LA2-AL-TR | SA1-CLT-SE | SA2-CLT-SE |

According to their theories, 'acquisition can take place only when people understand messages in the target language' and the input hypothesis maintains that in order for the learners to develop to the next stage in the acquisition of target language, they have to understand input language that includes a structure that is part of the next stage [11-36]. In other words, they should be given to study contents that are one stage ahead of their existing level of linguistic competence. Krashen refers to this with the formula "I + 1" (i.e., input that contains structures slightly above the learner's present level) [36] .  Why personalized experience is important is because it creates a stress-free environment, which is a fundamental need for language acquisition. According to Asher, acquisition takes place in a stress- free environment, whereas the learning environment often causes stress and anxiety [127].  Hence, to transform the process of 'learning' a language into a process of 'acquisition', creating a stress-free environment is vital as well.

Each of the cluster domains under the zone of proximal development contains two different cluster domains, one containing the skill-wise existing knowledge potential of the EFL learners and the one next to it contains the materials to achieve their potential knowledge zone in the skill. Data scientist may have to apply multiple algorithms such as K-Nearest Neighbours (KNN), Naive Bayes, Support vector Machines., etc. The materials in the cluster domain of potential knowledge zone link to contents that are one stage ahead of their current knowledge level in the module. For example, the student bearing the ID number BB6, likes to read texts on Travel and Adventure and his teacher prefers Task-Based learning contents as his personalized choice. His current level attribute in reading skill corresponds to A2 in CEFR scale. So, the teacher chooses for him contents from B1, which is designed with materials that are one stage ahead of his current competence level in reading. As soon as the teacher clicks on the cluster domain hyperlinked as 'RB1-TBL-TA under Potential Knowledge Zone, machine learning algorithm will direct him to the contents that are prepared following the modalities, tasks and activities of Task Based Learning by using travel and adventure texts only.

## 7.7. Sustainable Operation: Smart transformation of the Institution

The cluster domains, populated with pre-determined personalized, skill-wise and attribute-wise EFL contents, enable the machine learning algorithms [155]  to cull out patterns to identify personalized preferences, modalities, regression and differences existing in each domain. Machine learning algorithms can apply this knowledge of patterns without being supervised in open big data platform, transforming each cluster domain with huge volumes of structured data that can be used for EFL teaching and learning. Unsupervised learning happens when machine learning builds clusters from unknown data from another or same platform [159]. For example, the contents stored in reading domain carrying the hyperlinks RA2-DM-TR are mostly expressions in simple structures, and their modalities, activities and tasks resonate with Direct method of teaching and learning approach and these contents are based on technology and research field. Likewise, each of the domains has their own patterns, modalities, tasks, contexts and specific themes. According to machine learning theories, if the machine learning algorithm is able to identify the patterns prevailing in them, they can apply the same patterns in open bigdata platform to perform new tasks to store similar contents in the same domains [156]. The data scientists and the EFL teachers can evaluate the quality of the data and work further if there are anomalies or gross dissimilarities.

However, it's not necessary that the technologies or software that are vital to achieve, maintain and harness the benefits of Big Data in all these areas have to be self-invented, they can be borrowed, purchased or rented as well from the allies as well. For processing large scale data and machine learning, Apache Spark can be used as a unified analytics engine.  The task of determining appropriate technologies lies on the data scientists. It is part of his job to decide the types of data processing platform the institute may need for building EFL bigdata ecosystem. According to Dr. Mahmudur Rashid, "Intelligent analyst's ways of thinking in probabilities can help the organization find out new data insight, strengthening the efforts to transform learning into acquisition for the students". The EFL teachers need not bother about the technological integration process, as their job, in this connection, is to develop contents and guide the data scientists about the types of contents required for each cluster domains. Again, the job of content development can be done by writers and content developers as well.

The support services range from ethics to training; from classroom management to data management; from operational knowhow to evaluation of the performance; from pedagogical methods to students' cognitivism; from dealing with a class of students with mixed academic abilities to ensuring better understanding of the learners; form teaching best practices to professional challenges and so on. Dr. Rashid thinks that the interoperability will be difficult if all the departments, or colleges or EFL faculties fail in coordination among themselves and with the data scientists. "In addition to teaching", Dr. Rashid continues, "the EFL teachers have to work as the domain experts to guide the data scientists so that they can operate the system to achieve sustainable advantages." The leaders upfront, having determined the ethical and policy framework, should lead and set targets and goals for each of the EFL teaching faculties, colleges, or departments. The ongoing hype to digitalize education system should not be marred by uncertainties and lack of clear policies.

## 8. The Challenges

Often the old or historical unstructured contents are in image format, not easily readable. To be able to extract contents out of it, advanced Optical Character Recognition (OCR) techniques are required [180]. Moreover, unstructured data requires advanced techniques for feature extraction and representation, as traditional methods designed for structured data may not be directly applicable. Another challenge is the sheer volume of unstructured data, which can be massive and require specialized storage and processing capabilities. Lastly, unstructured data analysis often requires domain-specific knowledge and expertise to interpret and derive insights effectively. Overcoming these challenges involves employing techniques such as natural language processing, computer vision, text mining, and deep learning to handle unstructured data effectively and extract valuable information from it.

## 9. Conclusion

The integration of disruptive technologies, such as big data and analytics and machine learning can potentially contribute to facilitate smart transformation of EFL teaching and learning approaches, offering the stakeholders a personalized experience in EFL teaching and learning. The widespread use of internet, social media, smart phone and other smart devices such as smart boards, notepads or laptops objectively justifies that the many stakeholders carry the basic technological knowhow to cope up with such paradigm shift. The collaborative partnership building with the data scientists and EFL teachers will ease the process of building EFL Bigdata Ecosystem, enabling the stakeholders to develop customized EFL teaching and learning materials. Regardless of the differences in teaching methods and strategies followed by the EFL teachers in the classroom, the incorporation of smart strategy has the catalytic prospects to make learning more engaging and interesting, transforming the process of 'learning' into a process of 'acquisition' and facilitating a smart transformation of the commonly used teaching methods. The contents of the cluster domain of RA2-DM-AC are one stage ahead of the contents of RA2-DM-AC, which clearly creates an environment to measure the progress scientifically. The Zone of Proximity Development carries the mechanism to identify the ways the students learn and their progress during the courses. It is something unique, as it needs no extra efforts. In other words, it will help identify the trajectory of learning. The traditional system of evaluation of the students relies on their academic grades. But the Zone of Proximal Development will empower the teachers to evaluate the learning process, making the approach scientifically reliable. Further to that, offering a personalized experience in learning creates a stress-free environment. The contents of each cluster domain of EFL bigdata ecosystem are designed to engage the students. They generate an impactful appeal to their subconscious, inspire their cognitive process of learning, link their personal and social experience and engage them spontaneously with the situations and contexts. Leveraging upon fruitful interactions and upon customized teaching environment that are tailored to the diverse needs of the teachers and the learners, the approach creates a unique and smart platform to achieve their teaching and learning goals.

**Acknowledgement**

The author acknowledges the insightful observations Dr. Mohammad Mahmudur Rashid, Senior Data and Infrastructure Engineer at Schlumberger, Qatar, Dr. Shiladitya Sen, Montclair State University, NY, America and Professor Tania Hossain, Waseda University, Japan.

**References**

[1]     Cindy E. Hmelo-Silver, "Problem-based learning: What and how do students learn?", *Educational Psychology Review*, Vol. 16, No. 3, pp. 235-266, 2002, Print ISSN: 1040-726X, Online ISSN: 1573-336X, DOI: 10.1023/B: EDPR.0000034022.16470.f3.

[2]     George D. Kuh, "Assessing what really matters to student learning: Inside the National Survey of Student Engagement", *Change*, Vol. 33, No. 3, pp. 10-7, 2001, ISSN: 00091383, DOI: 10.1080/00091380109601795.

[3]     Kenneth A. Feldman, "What matters in college? Four Critical Years Revisited", *The Journal of Higher Education*, Vol. 65, No. 5, pp. 615–622, Print ISSN: 0022-1546, Online ISSN: 1538-4640, 1994, Available: https://doi.org/10.2307/2943781.

[4]     Ian Douglas and Nicole D. Alemanne, "Measuring student participation and effort", In *proceeding of the IADIS International Conference on Cognition and Exploratory Learning in Digital Age (CELDA)*, Algarve, Portugal, 7-9 December 2007, ISBN: 978-972-8924-48-5, Available: https://www.iadisportal.org/celda-2007-proceedings.

[5]     Ernest T. Pascarella and Patrick T. Terenzini, "How College Affects Students: Findings and Insights form Twenty Years of Research", *Academe*, Vol. 78, No. 4, pp. 46-47, Print ISSN: 2153-8492, 1991, DOI: https://doi.org/10.2307/40250363.

[6]     Anne Jelfs, Roberta Nathan and Clive Barrett, "Scaffolding students: Suggestions on how to equip students with the necessary skills for studying in a blended learning environment", *Journal of Educational Media*, Vol. 29, No. 2, pp. 85-95, 2004, Available: https://doi.org/10.1080/1358165042000253267.

[7]     Paul Ginns, "Quality in blended learning: Exploring the relationships between on-line and face-to-face teaching and learning", *Internet and Higher Education*, Vol. 10, No. 1, pp. 53-64, Online ISSN: 1873-5525, Print ISSN: 1096-7516, 2007, DOI: 10.1016/j.iheduc.2006.10.003.

[8]     Charles R. Greenwood, Betty T. Horton and Cheryl A. Utley, "Academic engagement: Current perspective in research and practice", *School Psychology Review*, Vol. 31, No. 3, pp. 328-349, 2002, Available: https://doi.org/10.1080/02796015.2002.12086159.

[9]     Marianne Perie and Rebecca Moran, "Three decades of student performance in reading and mathematics", *U.S. Department of Education Institute of Education Sciences*, Washington, DC, 2005, Available: https://nces.ed.gov/nationsreportcard/pubs/2005/2005464.asp.

[10]    Nettie Legters, Robert Balfanz and James McPartland, *Solutions for failing high schools: Converging visions and promising models*, Washington, DC, USA: Office of Vocational and Adult Education, 2002, Available: https://eric.ed.gov/?id=ED466942.

[11]    Stephen D. Krashen, *Second Language Acquisition and Second Language Learning*, Oxford, UK: Pergamon, 1981, Available: https://doi.org/10.1017/S0272263100004733.

[12]    Max S. Kirch, "Direct Method and the Audio-Lingual Approach", *The French Review*, Vol. 41, No. 3, pp. 383-385, 1967, Available: http://www.jstor.org/stable/385169.

[13]    N. S. Prabhu, "There is no best method-why?", *TESOL Quarterly*, Vol. 24, No. 2, pp. 161-176, Print ISSN: 0039-8322, Online ISSN: 1545-7249, 1990, Available: https://doi.org/10.2307/3586897.

[14]    Nina Garrett, "Technology in the service of language learning: Trends and Issues", *Modern Language Journal*, Vol. 93, pp. 74-101, Online ISSN: 1540-4781, 1991, DOI: 10.1111/j.1540-4781.2009.00968.x.

[15]    Bob Hoffman and Donn Ritchie, "Using multimedia to overcome the problems with problem-based learning", *Instructional Science*, Vol. 25, No. 2, pp. 97-115, Print ISSN: 0020-4277 Online Print ISSN: 1573-1952, 1997, Available: https://eric.ed.gov/?id=EJ546245.

[16]    M. Sharples, R. D. Roock, Rebecca Ferguson, Mark Gaved, C. Herodotou et al., "Innovating pedagogy 2016: Open University innovation report 5", *Institute of Educational Technology*, The Open University, 2016, Available: http://www.open.ac.uk/innovating.

[17]    Yin Yang, Yun Wen and Yanjie Song, "A Systematic review of technology-enhanced self-regulated language learning", *Educational Technology & Society*, Vol. 26, No. 1, pp. 31-44, Online ISSN: 1436-4522, Print ISSN: 1176-3647, January 2023, Available: https://www.j-ets.net/collection/published-issues/26_1.

[18]    J. B. Black, A. Segal, J. Vitale and C, Fadjo, "Embodied cognition and learning environment design" in *Theoretical foundations of student-centered learning environments*, D. J. &. S. Lamb, Eds., New York, Routledge, 2012, pp.198-223. Available: https://www.tc.columbia.edu/faculty/jbb21/faculty-profile/files/8Embodied_Cognition2.pdf.

[19] Chih-Wei Chang, Jih-Hsien Lee, Po-Yao Chao, Chin-Yeh Wang and Gwo-Dong Chen, "Exploring the Possibility of Using Humanoid Robots as Instructional Tools for Teaching a Second Language in Primary School", *Journal of Educational Technology & Society*, Vol. 13, No. 2, pp. 13-24, Print ISSN: 1176-3647, Online ISSN:1436-4522, 2010.

[20] Frank Klassner, "A case study of Lego Mindstorms' suitability for artificial intelligence and robotics course at the college level", in *Proceedings of the 33rd SIGCSE Technical Symposium on Computer Science Education*, New York, USA, 2002, Available: https://doi.org/10.1145/563517.563345.

[21] HyeJin Ryu, Sonya S. Kwak and Myung Suk Kim, "A Study on External Form Design Factors for Robots as Elementary School Teaching Assistants", *The Bulletin of Japanese Society for Science of Design (JSSD)*, Vol. 54, No. 5, 2008, DOI: 10.1109/ROMAN.2007.4415236.

[22] Sieglinde Jornitz, Laura Engel, Bernard Veldkamp, Kim Schildkamp, Merel Keijsers *et al.*, "Big Data Analytics in Education: Big Challenges and Big Opportunities", in *International Perspectives on School Settings, Education Policy and Digital Strategies*, pp. 266-282, Verlag Barbara Budrich, 2021, Available: https://doi.org/10.2307/j.ctv1gbrzf4.19.

[23] Zacharoula Papamitsiou and Anastasios A. Economides, "Learning Analytics and Educational Data Mining in Practice: A Systematic Literature Review of Empirical Evidence", *Journal of Educational Technology & Society*, Vol. 17, No. 4, pp. 49-64, 2014, Online ISSN: 1436-4522, Print ISSN: 1176-3647.

[24] Sylviane Granger, Olivier Kraif, Claude Ponton, Georges Antoniadis and Virginie Zampa, "Integrating learner corpora and natural language processing: A crucial step towards reconciling technological sophistication and pedagogical effectiveness", *ReCALL Journal*, Vol. 19, No. 3, pp. 252-268, Print ISSN: 0958-3440, Online ISSN: 1474-0109, 2007, DOI: 10.1017/S0958344007000237.

[25] David M Bell, "Do teachers think that methods are dead?", *ELT Journal*, Vol. 61, Online ISSN: 1477-4526, Print ISSN: 0951-0893, 2007, Available: https://doi.org/10.1093/elt/ccm006.

[26] Rod Ellis, *Task-based language learning and teaching*, Oxford, England: Oxford University Press, 2003, DOI: https://doi.org/10.1017/S0272263104293056.

[27] Scott Thornbury, "Why I wrote 30 Language Teaching Methods", *Cambridge English Blog*, Cambridge, September 2020, Available: https://www.cambridge.org/elt/blog/2020/09/29/why-i-wrote-30-language-teaching-methods/.

[28] Eli Hinkel, "Current Perspectives on Teaching the Four Skills", *TESOL Quarterly*, Vol. 40, No. 1, pp. 109-131, Print ISSN: 0039-8322, Online ISSN: 1545-7249, 2006, Available: https://www.jstor.org/stable/40264513.

[29] H. Douglas Brown and Heekyeong Lee, *Teaching by principles: an interactive approach to language pedagogy*, New York, USA: Pearson Education, 2015.

[30] B. Kumaravadivelu, "TESOL Methods: Changing Tracks, Challenging Trends", *TESOL Quarterly*, Vol. 40, No. 1, pp. 59-81, Print ISSN:0039-8322, Online ISSN: 1545-7249, 2006, Available: https://doi.org/10.2307/40264511.

[31] Jack C. Richards and Theodore Rodgers, *Approaches and Methods in Language Teaching*, Cambridge, UK: Cambridge University Press, 2001. DOI: https://doi.org/10.1017/CBO9780511667305.

[32] Gertrude Moskowitz, *Caring and Sharing in the Foreign Language Classroom*, Boston, USA: Heinle and Heinle, ISBN-10: 0838427715, 1978, Available: https://doi.org/10.1016/0346-251X(80)90013-5.

[33] Jane Spiro, "Learning Theories and Methods", in *Changing Methodologies in TESOL*, Edinburgh, Edinburgh University Press, 2013, pp. 11-34.

[34] Xieling Chen, Di Zou, Haoran Xie and Gary Cheng, "Twenty Years of Personalized Language Learning: Topic Modeling and Knowledge Mapping", *Educational Technology & Society*, Vol. 24, No. 1, pp. 205-222, Online ISSN: 1436-4522 Online and 1176-3647 Print, 2021, Available: https://www.jstor.org/stable/26977868.

[35] Jack C. Richards and Theodore Rodgers, "Natural Approach", in *Approaches and Methods in Language Teaching*, Cambridge, Cambridge Language Teaching Library, p. 132, ISBN-13: 978-0-521-00843-3, 2001, Available: https://assets.cambridge.org/97805218/03656/frontmatter/9780521803656_frontmatter.pdf.

[36] D. Krashen and Tracy D. Terrell, *The Natura l Approach: Language Acquisition in the Classroom*, Oxford, UK: Pergamon, 1983, Available: https://doi.org/10.1017/S0272263100005659.

[37] Mike Levy, "Technologies in Use for Second Language Learning", *The Modern Language Journal*, Vol. 93, pp. 769-782, Print ISSN: 0026-7902, Online ISSN: 1540-4781, 2009, Available: https://doi.org/10.1111/j.1540-4781.2009.00972.x.

[38] B. Kumaravadivelu, *Understanding language teaching: From method to post method*, New Jersey, USA: Lawrence Erlbaum, 2006, DOI: 10.4324/9781410615725.

[39] Dick Allwright, "The Death of the Method: Plenary Paper for the SGAV Conference, Carleton University, Ottawa, May 1991", *Centre for Research in Language Education Lancaster: Working papers No. 10*, Volume 10, 1991, Lancaster, England, The University of Lancaster.

[40] H. Douglas Brown, "English language teaching in the 'post-method' era: Towards better diagnosis, treatment, and assessment", in *Methodology in language teaching*, Jack C Richards and Willy A. Renandya, Eds., Cambridge, England, Cambridge University Press, 2002, pp. 9-18. Available: https://doi.org/10.1017/CBO9780511667190.003.

[41] David Nunan, *Designing tasks for communicative classroom*, Cambridge, England: Cambridge University Press, 1989, Available: https://doi.org/10.1017/S0272263100009578.

[42] Julia Lane, "The Role of Education and Training", *Journal of Policy Analysis and Management*, Vol. 35, No. 3, Online ISSN: 1520-6688, 2016, Available: https://doi.org/10.1002/pam.21922.

[43] Jo Handelsman, Diane Ebert-May, Robert Beichner, Peter Burns, Amy Chang et al., "Scientific teaching", Science, Vol. 521, p. 304, Print ISSN: 0036-8075, Online ISSN: 1095-9203, 2004, DOI: 10.1126/science.1096022.

[44] Kitchin, R., The data revolution. Big data, open data, data infrastructures and their consequences, Los Angeles: SAGE, 2014, Available: https://doi.org/10.4135/9781473909472.

[45] Matthew Wall, "Big data: are you ready for blast-off?", BBC News, 4 March 2014, Available: https://www.bbc.com/news/business-26383058.

[46] Shona McCombes and Tegan George, "What is Research Methodology? Steps and Tips", Scriber, January 2023, Available: https://www.scribbr.com/dissertation/methodology.

[47] B. Ekstrom Denemark, L. M. Jacobsen and J. C. Karlson, Explaining society critical realism in the social science, London, UK: Routledge, 2002.

[48] K. Stenius, K. Mäkelä, M. Miovský and R. Gabrhelík, "How to Write Publishable Qualitative Research" in *Publishing Addiction Science*, Ubiquity Press, pp. 155-172, 2017, DOI: 10.5334/bbd.h.

[49] Annika Wilmers and Sieglinde Jornitz, *International Perspectives on School Settings, Education Policy and Digital Strategies: A Transatlantic Discourse in Education Research*, Berlin, Germany: Verlag Barbara Budrich, pp. 266-282, 2021, DOI: 10.25656/01:21720.

[50] Ijeoma Onyeji-Nwogu, M. Bazilian and Todd J. Moss, "The Digital Transformation and Disruptive Technologies: Challenges and Solutions for the Electricity Sector in African Markets", *Center for Global Development*, 2017.

[51] L. Daigle, "On the Nature of the Internet", In *A Universal Internet in a Bordered World, Centre for International Governance Innovation and International Development Research Centre (IDRC)*, Waterloo, Ontario, Canada,, 2016, Available: https://www.cigionline.org/sites/default/files/documents/GCIG%20Volume%201_2.pdf.

[52] Kral Inge and Robert G. Schwab, "Learning Spaces: From the Local to the Global", in *Learning Spaces*, ANU Press, pp. 43-56, 2012.

[53] A. Richterich, "Big Data: Ethical Debates", in *The Big Data Agenda*, London, University of Westminster Press, 2018, pp. 33-52, Available: https://doi.org/10.2307/j.ctv5vddsw.

[54] Park Aarvik, "Big Data", in *Humanitarianism: Keywords*, Brill, 2020, pp. 10-12. Available: https://doi.org/10.1163/9789004431140_006.

[55] Ryan Burns and Jim Thacther, "What's so big about Big Data? Finding the spaces and perils of Big Data", *Geo Journal*, Vol. 80, No. 4, pp. 445-448, 2015, Available: https://doi.org/10.1007/s10708-014-9600-8.

[56] Katie Whipkey, "Guidance for Incorporating Big Data into Humanitarian", 2015, Available: http://digitalhumanitarians.com.

[57] M. Scholz, "Analytical Implementation", in *Big Data in Organizations and the Role of Human Resource Management*, Peter Lang AG, 2022, pp. 91-159, DOI: 10.3726/b10907.

[58] Maryam Alavi and Dorothy E. Leidner "AI Knowledge management and knowledge management systems: Conceptual foundations and research issues", *MIS Quarterly*, Vol. 25, No. 1, pp. 107-136, 2001, Print ISSN: 0276-7783, Online ISSN: 2162-9730, Available: https://dl.acm.org/doi/10.2307/3250961.

[59] Chaim Zins, "Conceptions of information science", *Journal of the American Society for Information Science and Technology*, Vol. 58, No. 3, pp. 330-350, 2007, Available: https://doi.org/10.1002/asi.20507.

[60] Marianna Ioannou and Andri Ioannou, "Technology-enhanced Embodied Learning", *Educational Technology & Society*, Vol. 23, No. 3, pp. 81-94, 2020.

[61] Maria Elena Corbeil, Joseph Rene Corbeil and Badrul H. Khan, "A Framework for Identifying and Analyzing Major Issues in Implementing Big Data and Data Analytics in E-Learning: Introduction to Special Issue on Big Data and Data Analytics", *Educational Technology*, Vol. 57, No. 1, pp. 3-9, 2017, Online ISSN: 0013-1962.

[62] Andrea De Mauro, Marco Greco and Michele Grimaldi, "A Formal Definition of Big Data Based on its Essential Features", *Library Review*, Vol. 65 No. 3, pp. 122-135, 2016, Available: https://doi.org/10.1108/LR-06-2015-0061.

[63] Danah Boyd and Kate Crawford, "Critical Questions for Big Data: Provocations for a Cultural, Technological, and Scholarly Phenomenon", *Information, Communication and Society*, Vol. 15, No. 5, p. 663, 2012, Print ISSN: 1468-4462, Online ISSN: 1369-118X, Available: https://doi.org/10.1080/1369118X.2012.678878.

[64] SAS, "What is Big Data?", SAS, 2013, Available: http://www.sas.com/en_be/insights/big-data/what-is-big-data.html.

[65] D. McCandless, "Data, information, knowledge, wisdom", 2010, Available: http://www.informationisbeautiful.net/2010/data-information-knowledge-wisdom.

[66] Ellen Rose, "Disrupting Disruptive Technology in Higher Education", Educational Technology, Vol. 54, No. 6, pp. 56-57, 2014, Online ISSN: 2476-0730.

[67] James Manyika, Michael Chui, Peter Bisson, Jonathan Woetzel and Richard Dobbs *et al.*, *Unlocking the potential of the Internet of Things*, New York, USA: Mckinsey Global Institute, 2015.

[68] Airbus, "Skywise", Airbus, 2022, Available: https://aircraft.airbus.com/en/services/enhance/skywise.

[69]  José Manuel Sánchez Losada, Javier Caamaño Eraso and Pilar Davila García, "Airport Management: The Survival of Small Airports", *International Journal of Transport Economics*, Vol. 39, No. 3, pp. 349-367, 2012, Online ISSN: 03918440.

[70]  Christopher P. Wright, Harry Groenevelt and Robert A. Shumsky, "Dynamic Revenue Management in Airline Alliances", *Transportation Science*, Vol. 44, No. 1, pp. 15-37, 2010, ISSN: 0041-16551.

[71]  Shawn McKay, Gavin S. Hartnett and Bruce Held, "Airline Security Through Artificial Intelligence: How the Transportation Security Administration Can Use Machine Learning to Improve the Electronic Baggage Screening Program", *Homeland Security Operational Analysis Center operated by the RAND Corporation*, 2022, Available: https://www.rand.org/pubs/perspectives/PEA731-1.html.

[72]  Jehn-Yih Wong and Pi-Heng Chung, "Retaining Passenger Loyalty through Data Mining: A Case Study of Taiwanese Airlines", *Transportation Journal*, Vol. 47, No. 1, pp. 17-29, ISSN: 00411612, 2008, Available: https://doi.org/10.2307/20713696.

[73]  B. Pesquet-Popescu, "Béatrice PESQUET-POPESCU Research and Innovation Director for the Surface Radar business", *Thales Group*, 2022, Available: https://www.thalesgroup.com/en/speakers-bureau/beatrice-pesquet-popescu.

[74]  danah boyd and Kate Crawford, "Critical Questions for Big Data: Provocations for a Cultural, Technological, and Scholarly Phenomenon", *Information, Communication & Society*, Vol. 15, No. 5 , pp. 662-679, 2012, Print ISSN: 1468-4462, Online ISSN: 1369-118X,  Available: https://doi.org/10.1080/1369118X.2012.678878.

[75]  The White House, "The President's Council of Advisors in Science and Technology Report", Washigton D.C., Available: https://www.whitehouse.gov/pcast/.

[76]  Kei Koizumi,  "Investing in America's Future through R&D, Innovation, and STEM Education: The President's FY 2016 Budget", *White House Office of Science and Technology*, Washington, D.C, 2 February 2015, Available: https://obamawhitehouse.archives.gov/blog/2015/02/02/investing-america-s-future-through-rd-innovation-and-stem-education-president-s-fy-2.

[77]  C. Dede, "Next Steps for "Big Data" in Education: Utilizing Data-Intensive Research", *Educational Technology*, Vol. 56, No. 2, pp. 37-42, 2016, Available: https://dash.harvard.edu/handle/1/28265473.

[78]  Cristobal Romero and Sebastian Ventura, "Data mining in education", *Wiley Interdisciplinary Reviews: Data Mining and Knowledge Discovery*, Vol. 3, No. 1, pp. 12-27, 2013, Available: https://doi.org/10.1002/widm.1075.

[79]  Frank Robbin te Velde and Cor-Jan Jager, "Big Data in education and science. Inventory and essays", Dialogic, 2014, Available: https://www.dialogic.nl/en/?s=Big+Data+in+education+and+science.+Inventory+and+essays.

[80]  James G. Holland and B. F. Skinner, "The analysis of behavior: A program for self-instruction", *MULL: Modern Uses of Logic in Law*, Vol. 3, No. 3, pp. 139-144, 1962.

[81]  Thomas Ohse, "Teaching English in India", *ResearchGate*, p. 1, 2017, DOI: 10.13140/RG.2.2.27775.82080, Available: https://www.researchgate.net/publication/309014508_Teaching_English_in_India.

[82]  Anjana Rajeev T., "Methods and Approaches of Teaching English Language in India: An Eclectic Overview", *GAP BODHI TARU, A Global Journal of Humanities*, Vol. 3, No. 4, p. 34, 2020, Available: https://archive.org/details/34-38-methods-and-approaches-of-teaching-english-language-in-india-an-eclectic-overview.

[83]  Mohammad Mosiur Rahman and Ambigapathy Pandian, "A critical investigation of English language teaching in Bangladesh: Unfulfilled expectations after two decades of communicative language teaching. English Today", *English Today*, Vol. 34, No. 3, p. 43–49, 2018, DOI: 10.1017/S026607841700061X.

[84]  Zhenhui Rao and Chunhua Lei, *Teaching English as a foreign language in Chinese universities: The present and future*, Cambridge, UK: Cambridge University Press, 2014, Available: https://doi.org/10.1017/S026607841400039X.

[85]  Yanrini Martha Anabokay, "TEFL Methods in Indonesia", *International Journal of Linguistics Literature and Culture*, Vol. 5, No. 2, pp. 13-23, 2019, DOI: 10.21744/ijllc.v5n2.612.

[86]  Diane Larsen-Freeman, *Techniques and Principles in Language Teaching*, Oxford, UK: Oxford University Press, 1986.

[87]  Braj B. Kachru, "World Englishes: Agony and Ecstasy", *Journal of Aesthetic Education*, Vol. 30, No. 2, pp. 135-155, 1996, Online ISSN: 1543-7809, Print ISSN: 0021-8510, Available: https://doi.org/10.2307/3333196.

[88]  Amy Dayton-Wood, "Teaching English for A Better America", *Rhetoric Review*, Vol. 27, No. 4, pp. 397-414, 2008.

[89]  S. D. Janjuha-Jivraj, "Harnessing The Power of Diversity Through Pluralism", *Forbes*, 10 July 2019, Available: https://www.forbes.com/sites/shaheenajanjuhajivrajeurope/2019/07/10/harnessing-the-power-of-diversity-through-pluaralism/?sh=4feed973729b.

[90]  Sandra J. Savignon, *Communicative competence: Theory and classroom practice*, R. MA and Addison-Wesely, Eds., New York, USA: McGraw-Hill, 1997, ISBN: 9780070837362.

[91]  David Numan, *Designing tasks for the communicative classroom*, Cambridge, England: Cambridge University Press, 1989.

[92]  M. P. Breen and C. N. Candlin, "The essentials of a communicative curriculum in language teaching", *Applied Linguistics*, Vol. 1, pp. 89-110, 1980.

[93]   Michael A Canale and Merrill Swain, "Theoretical bases of communicative approaches to second language teaching and testing", *Applied Linguistics*, Vol. 1, pp. 1-47, 1980, Print ISSN: 0142-6001, DOI:10.1093/applin/I.1.1.

[94]   J. L. Austin, *How to do things with words*, London, UK: Oxford University Press , 1962.

[95]   Dell Hymes, "On communicative competence", in *Sociolinguistics: Selected readings*, J. P. and J. Holmes, Eds., Harmondswort, Penguin Books, pp. 269-293, 1972.

[96]   Johanna Kawasaki, "5 Popular ESL Teaching Methods Every Teacher Should Know", *Bridge Education Group*, 8 December 2021, Available: https://bridge.edu/tefl/blog/esl-teaching-methods/.

[97]   Michael Legutke and Howard Thomas, *Process and experience in the language classroom*, London, UK: Longman, 1991.

[98]   Scott Thornbury, "Teachers research teacher talk", *English Language Teaching Journal*, Vol. 50, No. 4, pp. 279-288, 1996, Available: https://doi.org/10.1093/elt/50.4.279.

[99]   Henry Widdowson, *Defining issues in English language teaching*, Oxford, England: Oxford University Press, 2003.

[100]  Michael Swan, "A critical look at the communicative approch", *English Language Teaching Journal*, Vol. 39, No. , pp. 2-12, Online ISSN: 1477-4526, 1985, Available: https://doi.org/10.1093/elt/39.1.2.

[101]  Stephen Evans, "The context of English language education", *RELC Journal*, Vol. 27, No. 2, pp. 30-55, 1996, Available: https://doi.org/10.1177/003368829602700203.

[102]  Cecilia Lai, "Communication failure in the language classroom: An explanation of causes", City University of Hong Kong, Hong Kong SAR, China, 1993, Available: https://doi.org/10.1177/0033688294025001.

[103]  Tomlinson, Brain, "Conflicts in TEFL: Reasons for failure in secondary schools", *Institute of Language in Education*, Vol. 4, pp. 103-110, 1988, Available: https://bibliography.lib.eduhk.hk/bibs/22b6cae4.

[104]  R. Mitchell, *Communicative Language Teaching in Practice*, London, UK: Centre for Information on Language Teaching and Research, pp. 23-24, 1988.

[105]  Klaus K. Brandl, *Communicative Language Teaching in Action: Putting Principles to Work*, New Work: Pearson Prentice Hall, 2007, ISBN: 9780131579064.

[106]  Jack C. Richards, *Communicative Language Teaching Today*, Cambridge, UK: Cambridge University Press, 2006, pp. 14–21.

[107]  Andrew Reimann, "Behaviorist Learning Theory", *The TESOL Encyclopedia of English Language Teaching*, pp. 1–6, 2018, Available: https://doi.org/10.1002/9781118784235.eelt0155.

[108]  Diane Larsen-Freeman, *Techniques and Principles in Language Teaching*, Oxford, UK: Oxford University Press, 2000, ISBN: 9780194355742200.

[109]  Jeremy Harmer, *The Practice of English Language Teaching*, 5th ed., Essex, UK: Pearson Education Ltd., 2001, ISBN: 9781447980254.

[110]  Harold Byron Allen, Russell N. Campbell, *Teaching English as a Second Language*, 2nd ed., New York, USA: McGowan-Hill, pp. 98-100, 1972, ISBN: 9780070010710.

[111]  Rabea Ali, "A Review Of Direct Method And Audio-Lingual Method In English Oral Communication", *International Journal of Scientific & Technology Research*, Vol. 9. No. 8, ISSN: 2277-861, 2020.

[112]  Noam Chomsky, Reflections on Language, New York, USA: Pantheon Books, 1975, ISBN: 9780394499567.

[113]  K Byrappa, *Content Cum Methodology of Teaching English*, Banglore, India: Sapna Book house, 2013, ISBN-13: 978-8128010682.

[114]  Babu Muthuja, *Teaching of English*, South Africa: Centurium Press, p.87, 2009, ISBN-13: 978-9380252407.

[115]  Inlingua, "2023 inlingua Galindo Beograd", *Inlingua*, 2023, Available: https://www.inlingua-beograd.com/sr/akcije-i-popusti.

[116]  Y. K. Singh, *Teaching of English*, Banglore, India: APH publishing corporation, p. 66, 2005, ISBN-10: 8176487112, ISBN-13: 9788176487115.

[117]  Nguyễn Nhật Quang, Phạm Nhật Linh and Nguyễn Thị Thu Hiền, "Tasks, self-efficacy, and L2 motivational self system in an online emergency EFL speaking class: A mixed-methods study", *The JALT CALL Journal*, Vol. 18, No. 1 , pp. 1-33, 2022, ISSN: 1832-4215, Available: https://doi.org/10.29140/jaltcall.v18n1.518.

[118]  N. S. Prabhu, *Second Language Pedagogy*, Oxford, UK: Oxford University Press, 1987, ISBN: 9780194370844.

[119]  Susanne Niemeie, *Task-based grammar teaching of english: Where cognitive grammar and task-based language teaching meet*, Tübingen, Germany: A. Francke Verlag, 2017, ISBN: 9783823391302.

[120]  Betty Lou Leaver and Jane R. Willis, *Task-Based Instruction in Foreign Language Education: Practices and Programs*, Georgetown, USA: Georgetown University Press, 2004, ISBN-13: 978-1589010284.

[121]  J. Larsson, "Problem-Based Learning: A possible approach to language education?", *Polonia Institute*, Jagiellonian University, 2001.

[122]  D. Carless, "Issues in Teachers' Reinterpretation of a Task-Based Innovation in Primary Schools", *TESOL Quarterly*, Vol. 38, No. 4 , pp. 639-662, 2004, Available: https://doi.org/10.2307/3588283.

[123]  Merrill Swain and Sharon Lapkin, "Task-based second language learning: The uses of the first language", *Language Teaching Research*, Vol. 4, No. 3, pp. 251-274, 2000, Online ISSN: 1477-0954, Available: https://doi.org/10.1177/136216880000400304.

[124] A. B. M. Tsui, *Understanding expertise in teaching: Case studies of ESL Teachers*, Cambridge, England: Cambridge University Press, 2003.

[125] Paul Seedhouse, "Task-based interaction", *ELT Journal*, Vol. 53, No. 3, pp. 149-156, 1999, Available: https://doi.org/10.1093/elt/53.3.149.

[126] Richard Frost, "A Task-based Approach", *British Council Teaching English*, Turkey, 2015, Available: https://www.teachingenglish.org.uk/professional-development/teachers/knowing-subject/articles/task-based-approach.

[127] James J. Asher, *Learning Another Language Through Actions*, 5th ed., California, USA: Sky Oak Productions, 1996.

[128] Paula Conroy, M. A., "Total Physical Response: An Instructional Strategy for Second-Language Learners who are Visually Impaired", *SAGE Journals*, Vol. 93, No. 5, 2019, ISSN: 2158-2440. DOI: https://doi.org/10.1177/0145482X990930050.

[129] L. Zink de Diaz, "TPR Foreign Language Instruction and Dyslexia", *Davis Dyslexia Association International*, 2005, Available: https://www.dyslexia.com/about-dyslexia/understanding-dyslexia/tpr-foreign-language-instruction/.

[130] Michael Byram, *Routledge Encyclopedia of Language Teaching and Learning*, London, UK: Routledge, 2000, pp. 631-633, Available: https://doi.org/10.4324/9780203219300.

[131] Vivian Cook, *Second Language Learning and Language Teaching*, 5th ed., London, UK: Hodder Education, 2016, Available: https://doi.org/10.4324/9781315883113.

[132] Aandy Kirkpatrick, "World Englishes: Implications for International Communication and English Language Teaching", *Language in Society*, Volume 38, No. 4, September 2009, pp. 537-538, 2007, Available: https://doi.org/10.1017/S0047404509990376.

[133] Haixiao Wang and Clifford Hill, "A Paradigm Shift for English Language Teaching in Asia: From Imposition to Accommodation", *The Journal of ASIA TEFL*, Vol. 8, No. 4, pp. 205-230, 2011.

[134] C. Pichler, "Language Learning through the Eye and Ear Webcast", *Gallaudet University*, 2022, Available: https://clerccenter.gallaudet.edu/national-resources/learning/learning-opportunities/webcasts/language-learning-through-the-eye-and-ear-webcast.html.

[135] Stephen Krashen, "ESL Glossary", *Boogles World ESL*, 2022, Available: https://bogglesworldesl.com/glossary/acquisitionlearninghypothesis.htm.

[136] H. U. Brown, *First Language: The Early Stages*, Cambridge, UK: Harvard University Press, 1973.

[137] Ulla Connor, *Contrastive Rhetoric: Cross-cultural aspects of second-language writing*, Cambridge, UK: Cambridge University Press, ISBN: 9780521446884, 1996.

[138] Robert Lado and Ann Arbor, "Linguistics Across Cultures: Applied Linguistics for Langauge Teachers", *Canadian Journal of Linguistics/Revue canadienne de linguistique*, Vol. 3, No. 2, October 1957, pp. 85 - 87, Available: https://doi.org/10.1017/S0008413100025196.

[139] Michael Swan and Bernard Smith, "Learner English: A Teacher's guide to interference and othe problems", *Studies in Second Language Acquisition*, Vol. 10, No. 3, October 1988, pp. 406-407, 2001, Available: https://doi.org/10.1017/S0272263100007555.

[140] L. S. Vygotsky and Michael Cole, *Mind in society: Development of Higher Psycholgoical Processes*, Cambridge, UK: Harvard University Press, 1978, ISBN: 9780674576292.

[141] J. Coleman, "Make Learning a Lifelong Habit", *Harvard Business Review*, 24 January 2017, Available: https://hbr.org/2017/01/make-learning-a-lifelong-habit.

[142] WGU, "What Humanistic Learning Theory", *Western Governors University*, 21 July 2020, Available: https://www.wgu.edu/blog/what-humanistic-learning-theory-education2007.html#close.

[143] Francisco Gomes de Matos, "Caring and sharing in the foreign language class: Moskowitz, Gertrude, A Sourcebook on Humanistic Techniques. Rowley, Mass", *System*, Vol. 8, No. 3, October 1980, p. 274-276, 1978, Available: https://doi.org/10.1016/0346-251X(80)90013-5.

[144] J. Arnold, *Affect in Language Learning*, Cambridge, UK: Cambridge University Press, Cambridge: Cambridge University Press, 1999.

[145] CommerceMates, "Cognitive Learning Theory", *Commerce Mates*, 2018, Available: https://commercemates.com/cognitive-learning-theory/.

[146] Peggy A. Ertmer and Newby, "Behaviorism, Cognitivism, Constructivism: Comparing Critical Features from an Instructional Design Perspective", *Performance Improvement Quarterly*, Wiley, Vol. 6, No. 4, pp. 50-72, 1993.

[147] T. P. P. Center, "Cognitive Learning Theory", *The Peak Performance Center*, 2023, Available: https://thepeakperformancecenter.com/educational-learning/learning/theories/cognitive-learning-%20theory/.

[148] J. Rubin, "Learner strategies: theoretical assumptions, research history and typology", in *Learner Strategies in Language Learning*, A. W. A. J. Rubin, Ed., London, UK: Prentice Hall, 1987, pp. 15-30.

[149] R. Oxford, *Language Learning Strategies: What Every Teacher Should Know*, Boston, USA: Heinle and Heinle., 1990.

[150] R. Oxford and D. Crookall, "Research on Language Learning Strategies: Methods, Findings, and Instructional Issues", *The Modern Language Journal*, Vol. 73, No. 4, 1989.

[151] J. McKinley, "Critical Argument and Writer Identity: Social Constructivism as a Theoretical Framework for EFL Academic Writing", *Critical Inquiry in Language Studies*, Vol. 12, No. 3, pp. 184-207, 2015.

[152] C. Rayson, "Learner Autonomy: what you need to know", Cambridge, 21 January 2021, Available: https://www.cambridge.org/elt/blog/2021/01/29/learner-autonomy/.

[153] Fattene Zarrinkalam, Guangyuan Piao, Stefano Faralli and Ebrahim Bagheri "Mining User Interests from Social Media", in *Proceedings of the 29th ACM International Conference on Information & Knowledge Management (CIKM '20)*, New York, United States, October 2020, pp. 3519–3520, Available: https://doi.org/10.1145/3340531.3412167.

[154] U. O. Connecticut, "UCONN University of Connecticut", *The National Research Center on the Gifted and Talented*, 2013, Available: https://nrcgt.uconn.edu/underachievement_study/curriculum-compacting/cc_section11/.

[155] T. Mitchell, *Machine Learning*, New York, USA: McGraw Hill, 1997, ISBN: 0070428077, Available: https://www.cs.cmu.edu/~tom/mlbook.html.

[156] N. Luck, "Machine Learning Powered Artificial Intelligence" *Peace Research Institute Frankfurt*, Frankfurt, 2019, Available: https://www.hsfk.de/fileadmin/HSFK/hsfk_publikationen/prif0819.pdf.

[157] Shalev-Shwartz and Shai Ben-David, Understanding Machine Learning: From Theory to Algorithms, Cambridge, UK: Cambridge University Press, ISBN: 978-1-107-05713-5, 2014, Available: https://assets.cambridge.org/9781107/057135/frontmatter/9781107057135_frontmatter.pdf.

[158] IBM, "What is deep learning", *IBM*, 2022, Available: https://www.ibm.com/topics/deep-learning.

[159] S. J. Russell and P. Norvig, *Artificial Intelligence. A Modern Approach*, New Jersey, USA: Upper Saddle River, 2010, Available: http://aima.cs.berkeley.edu/.

[160] Vishmi Nayanathara, "What is Chat GPT & How Chat GPT works in 2023?", 2023, Available: https://digifix.com.au/what-is-chat-gpt-how-chat-gpt-works-in-2023/.

[161] W. J. Carbonaro and A. Gamoran, "The production of achievement inequality in high school English", *American Educational Journal*, Vol. 39, No. 4, pp. 801-827, 2002.

[162] A. Crosson, L. C. Matsumura, R. Correnti and A. Arlotta-Guerrero, "Writing tasks and students use of academic language", *The Elementary School Journal*, Vol. 112, No. 3, pp. 469-496, 2012.

[163] S. Graham and  M. A. Herbert, *Writing to read: Evidence for how writing can improve reading*, Wishington, DC, USA: Alliance for Excellent Education, 2010.

[164] I. C. Matsumra, H. Garnier, J. Pascal and R. Valde's, "Measuring Instructional quality in accountability systems: Classroom assignment and student achievement", *Educational Assessment*, Vol. 8, No. 3, pp. 207-229, 2002.

[165] L. C. Matsumura, H. Gamier, S. C. Slater and M. B. Boston, "Measuring instructional interactions 'at-scale'", *Educational Assessment*, Vol. 13, No. 4, pp. 267-300, 2008.

[166] W. Doyle and K. Carter, "Academic tasks in classrooms", *Curriculum Inquiry*, Vol. 14, No. 2, pp. 129-149, 1984.

[167] L. C. Matsumura, R. Correnti and E. Wang, "Classroom Writing Tasks and Students' Analytic Text-Based Writing", *Reading Research Quarterly*, Vol. 50, No. 4, pp. 417-438, 2015.

[168] Robbin Velde, Frank Bongers and Jager Cor-Jan, *Big Data in education and science. Inventory and essays*, Utercht: Dialogic, 2014.

[169] Paul Brabin and Andy Clayton, "Building your digital DNA", *Deloitte MCS Limited*, 2016, Available: http://www2.deloitte.com/uk/en/pages/technology/articles/building-your-digital-dna.html.

[170] Dean Walsh, "Technological Dystopia - 10 Reasons to Fear Technology", *Hubpages*, 2018, Available: https://discover.hubpages.com/technology/Technological-Dystopia-10-Reasons-to-Fear-Technology.

[171] U. Franke, "Harnessing Artificial Intelligence", *European Council on Foreign Relations*, 2019.

[172] N. Bastian, "Building the Army's Artificial Intelligence WorkforceAuthor", *The Cyber Defense Review*, Vol. 5, No. 2, pp. 59-64, 2020.

[173] Abdulaziz Aldiab, Harun Chowdhury, Alex Kootsookos, Firoz Alam and Hamed Allhibi, "Utilization of Learning Management Systems( LMS) in higher education system: A case review of Saudi Arabia", *Energy Procedia*, Vol. 160, pp. 731-737, February 2019, Available: https://www.sciencedirect.com/science/article/pii/S1876610219312767.

[174] L. Breiman, J. Friedman, R. Olshen and C. Stone, *Classification and Regression Trees*, Belmont, USA: Wardsworth, 1984.

[175] Tin Kam Ho, "Radom Decision Forests", in *Proceedings of the 3rd International Conference on Document Analysis and Recognition*, Montreal, Canada, 14-16 August 1995, Print ISBN: 0-8186-7128-9, DOI: 10.1109/ICDAR.1995.598994, Available: https://ieeexplore.ieee.org/document/598994.

[176] Asa Ben-Hur, David Horn, Hava T. Siegelmann and Vladimir Vapnik,  "Support Vector Clustering", *Journal of Machine Learning Research*, Vol. 2, pp. 125-137, ISSN: 1532-4435, 2001.

[177] David J. Hand and Keming Yu, "Idiot's Bayes - not so stupid after all?", *International Statistical Review*, Vol. 69, No. 3, pp. 385-399, 2001, Available: https://www.jstor.org/stable/1403452.

[178] T. M. Cover and P. E. Hart, "Nearest Neighbor Pattern Classification", *IEEE Transaction on Information Theory*, Vol. 13, pp. 21-27, 1967, Available:  https://isl.stanford.edu/~cover/papers/transIT/0021cove.pdf.

[179] Veena Ghorakavi, "Neural Networks | A beginners guide", 18 April 2023, Available: https://www.geeksforgeeks.org/neural-networks-a-beginners-guide/.

[180] Md. Russell Talukder, "Using sings and symbols in teaching English in global perspective", in *Proceedings of the Asian Conference on Language Learning 2013*, Osaka, Japan, 2013, pp. 551-566, Available: https://papers.iafor.org/submissionacll2013_0316/.

[181] E. E. Fournier d'Albe, "On a Type-Reading Optophone", in *Royal Society A: Mathematical, Physical and Engineering Sciences*, Vol. 90, No. 619, 1914, Print ISSN: 0950-1207, Online ISSN:2053-9150, Published by: Royal Society, Available: https://doi.org/10.1098/rspa.1914.0061.